\pdfoutput=1
\documentclass{article}

\usepackage{microtype}
\usepackage{graphicx}
\usepackage{subfigure}
\usepackage{booktabs} 

\usepackage{hyperref}

 \usepackage[accepted]{icml2024}

\usepackage{amsmath}
\usepackage{amssymb}
\usepackage{mathtools}
\usepackage{amsthm}

\usepackage[capitalize,noabbrev]{cleveref}

\theoremstyle{plain}
\newtheorem{theorem}{Theorem}[section]

\theoremstyle{definition}
\newtheorem{definition}[theorem]{Definition}

\theoremstyle{remark}

\usepackage[textsize=tiny]{todonotes}

\def\blue{\textcolor{blue}}
\def\purple{\textcolor{purple}}
\def\red{\textcolor{red}}
\def\yellow{\textcolor{yellow}}
\def\green{\textcolor{green}}

\def\a{{\bf a}}

\def\e{{\bf e}}

\def\mH{{\bf H}}

\def\I{{\bf I}}

\def\P{{\bf P}}

\def\S{{\bf S}}

\def\y{{\bf y}}
\def\z{{\bf z}}

\def\0{{\bf 0}}
\def\1{{\bf 1}}

\def\FM{{\mathcal F}}

\def\NM{{\mathcal N}}

\def\DM{{\mathcal D}}

\def\RB{{\mathbb R}}
\def\EB{{\mathbb E}}

\def\alp{\mbox{\boldmath$\alpha$\unboldmath}}

\def\Om{\mbox{\boldmath$\Omega$\unboldmath}}
\def\bet{\mbox{\boldmath$\beta$\unboldmath}}
\def\et{\mbox{\boldmath$\eta$\unboldmath}}

\def\ph{\mbox{\boldmath$\phi$\unboldmath}}

\def\tha{\mbox{\boldmath$\theta$\unboldmath}}

\def\muu{\mbox{\boldmath$\mu$\unboldmath}}
\def\Si{\mbox{\boldmath$\Sigma$\unboldmath}}

\def\Gam{\mbox{\boldmath$\Gamma$\unboldmath}}
\def\gamm{\mbox{\boldmath$\gamma$\unboldmath}}

\newcommand{\tabref}[1]{Table~\ref{#1}}
\newcommand{\secref}[1]{Sec.~\ref{#1}}
\newcommand{\figref}[1]{Fig.~\ref{#1}}

\newcommand{\thmref}[1]{Theorem~\ref{#1}}

\newcommand{\defref}[1]{Definition~\ref{#1}}
\newcommand{\eqnref}[1]{Eq.~\ref{#1}}
\newcommand{\algref}[1]{Alg.~\ref{#1}}
\newcommand{\appref}[1]{Appendix~\ref{#1}}

\renewcommand{\hat}{\widehat}
\renewcommand{\frac}{\tfrac}

\DeclareMathAlphabet{\mymathbb}{U}{bbold}{m}{n}
\definecolor{green}{rgb}{0,0.5,0}

\def\blue#1{\textcolor{blue}{#1}}

\def\red#1{\textcolor{red}{#1}}
\def\green#1{\textcolor{green}{#1}}
\def\purple#1{\textcolor{purple}{#1}}

\usepackage{amsmath}
\usepackage{amssymb}
\usepackage{mathtools}
\usepackage{amsthm}
\usepackage{wrapfig}
\usepackage{booktabs}
\usepackage{enumitem}
\renewcommand{\algref}[1]{Alg.~\ref{#1}}
\usepackage{multirow}
\usepackage[nice]{nicefrac}
\definecolor{cvprblue}{rgb}{0.21,0.49,0.74}
\definecolor{yellow}{rgb}{0.74,0.49,0}
\usepackage{enumitem}
\setitemize{noitemsep,topsep=0pt,parsep=0pt,partopsep=0pt}

\icmltitlerunning{Probabilistic Conceptual Explainers: Trustworthy Conceptual Explanations for Vision Foundation Models}

\begin{document}

\twocolumn[

\icmltitle{Probabilistic Conceptual Explainers:\\Trustworthy Conceptual Explanations for Vision Foundation Models}



\icmlsetsymbol{equal}{*}
\begin{icmlauthorlist}
\icmlauthor{Hengyi Wang}{equal,ru}
\icmlauthor{Shiwei Tan}{equal,ru}
\icmlauthor{Hao Wang}{ru}

\end{icmlauthorlist}

\icmlaffiliation{ru}{Department of Computer Science, Rutgers University, New Jersey, USA}

\icmlcorrespondingauthor{Hengyi Wang}{hengyi.wang@rutgers.edu}

\icmlkeywords{Machine Learning, ICML}

\vskip 0.3in
]



\printAffiliationsAndNotice{\icmlEqualContribution} 
\begin{abstract}

Vision transformers (ViTs) have emerged as a significant area of focus, particularly for their capacity to be jointly trained with large language models and to serve as robust vision foundation models. Yet, the development of trustworthy explanation methods for ViTs has lagged, particularly in the context of post-hoc interpretations of ViT predictions. Existing sub-image selection approaches, such as feature-attribution and conceptual models, fall short in this regard. This paper proposes five desiderata for explaining ViTs -- faithfulness, stability, sparsity, multi-level structure, and parsimony -- and demonstrates the inadequacy of current methods in meeting these criteria comprehensively. We introduce a variational Bayesian explanation framework, dubbed ProbAbilistic Concept Explainers (PACE), which models the distributions of patch embeddings to provide trustworthy post-hoc conceptual explanations. 
Our qualitative analysis reveals the distributions of patch-level concepts, elucidating the effectiveness of ViTs by modeling the joint distribution of patch embeddings and ViT's predictions. Moreover, these patch-level explanations bridge the gap between image-level and dataset-level explanations, thus completing the multi-level structure of PACE. 
Through extensive experiments on both synthetic and real-world datasets, we demonstrate that PACE surpasses state-of-the-art methods in terms of the defined desiderata\footnote{{Code will soon be available at https://github.com/Wang-ML-Lab/interpretable-foundation-models}}. 
\end{abstract}
\section{Introduction}
\label{sec:intro}

Vision Transformers (ViTs)~\citep{dosovitskiy2020vit} and their variants~\citep{liu2021swin,touvron2021deit,radford2021clip} have emerged as pivotal models in computer vision, leveraging stacked self-attention blocks to encode raw inputs and produce patch-wise embeddings as contextual representations. Given their increasing application in high-risk domains such as autonomous driving, explainability has become a critical concern.

To date, post-hoc explanations in computer vision often involve attributing predictions to specific image regions. However, we identify two primary limitations in current methods: (1) {
Existing conceptual explanation methods~\cite{fel2023holistic,fel2023craft,ghorbani2019ace,zhang2021ice,wang2020scorecam,novello2206making,chen2024less,li2021instance,wang2023learning} are not fully compatible with transformer-based models like vision transformers (ViTs), and they also fall short in offering a cohesive structure for dataset-image-patch analysis of input images.}
(2) Current state-of-the-art methods~\citep{li2020quantitative,pan2021agi,agarwal2022openxai,colin2022cannot,xie2022vit-cx,wang2022hint,fel2023craft,chen2023sim2word} evaluate visual concepts through subjective human utility scores or limited quantitative analysis, lacking a fair and consistent comparison framework. To address this, 
we propose a comprehensive set of desiderata for post-hoc conceptual explanations for ViTs,
namely (see formal definitions in~\secref{sec:concept_def}): 
\begin{itemize}[noitemsep,topsep=0pt,parsep=0pt,partopsep=0pt]
\item \emph{Faithfulness:} The explanation should be faithful to the explained ViT and able to recover its prediction. 
\item \emph{Stability:} The explanation should be stable for different perturbed versions of the same image. 
\item \emph{Sparsity:} For each prediction's explanation, only a small subset of concepts are relevant. 
\item \emph{Multi-Level Structure:} There should be dataset-level, image-level, and patch-level explanations. 
\item \emph{Parsimony:} There are a small number of concepts in total {(see~\appref{app:sparse_vs_pars} for more details)}. 
\end{itemize}
{While previous research~\cite{kim2018tcav,fel2023craft,oikarinen2023label, gilpin2018explaining, murdoch2019definitions} has proposed and met different dimensions of the learned concepts, these studies often lack a comprehensive evaluation. }
In this paper, propose ProbAbilistic Concept Explainers (PACE) to provide trustworthy conceptual explanations aligned with these desiderata, drawing inspiration from hierarchical Bayesian deep learning~\cite{BDL,BDLSurvey,NPN}. For example, to enable \textbf{\emph{multi-level}} explanations, we 
(1) model $K$ concepts as a mixture of $K$ Gaussian patch-embedding distributions, 
(2) treat the explained ViT's patch-level embeddings as observed variables, 
(3) learn a hierarchical Bayesian model that generates these embeddings in a top-down manner, from \emph{dataset-level} concepts through \emph{image-level} concepts to \emph{patch-level} embeddings, and (4) infer these multi-level concepts as our multi-level conceptual explanations; 
to enhance \textbf{\emph{stability}}, our hierarchical Bayesian model ensures that the inferred concepts from two different perturbed versions of the same image remain similar to each other. 
Our contributions are: 
\begin{enumerate}[noitemsep,topsep=0pt,parsep=0pt,partopsep=0pt]
    \item We comprehensively study a systematic set of five desiderata \emph{faithfulness, stability, sparsity, multi-level structure}, and \emph{parsimony} when generating trustworthy concept-level explanations for ViTs. 
    \item We develop the first general method,  dubbed ProbAbilistic Concept Explainers (PACE), as a variational Bayesian framework that satisfies these desiderata. 
    \item Through both quantitative and qualitative evaluations, our method demonstrates superior performance in explaining post-hoc ViT predictions via visual concepts, outperforming state-of-the-art methods across various synthetic and real-world datasets.
\end{enumerate}


\section{Related Work}
\textbf{Vision Transformers.} 
Vision Transformers (ViTs) have revolutionized computer vision by adapting the Transformer architecture for image recognition. The pioneering ViT model processes images as sequences of patches, surpassing traditional convolutional networks in efficiency and performance~\cite{dosovitskiy2020vit}. Subsequent innovations include the Swin Transformer~\cite{liu2021swin}, which introduces a hierarchical structure with shifted windows, and the Data-efficient Image Transformers (DeiT)~\cite{touvron2021deit}, which optimize training with a distillation token and teacher-student strategy. The CLIP model~\cite{radford2021clip} extends ViT's applicability by learning from natural language supervision, showcasing the architecture's versatility and robustness in visual representation learning.

\textbf{Visual Explanation Methods.} 
The landscape of visual {explanation methods~\cite{gilpin2018explaining, langer2021we, schwalbe2023comprehensive}} in computer vision is diverse, encompassing both feature attribution and concept-based approaches. Prominent methods such as LIME and SHAP~\cite{ribeiro2016lime, lundberg2017shap,simonyan2013saliency,li2021deeplift,shrikumar2017learning} provide insights by assigning importance scores to input features, enhancing the understanding of model decisions. Alongside these, concept-based explanations are also gaining popularity. \emph{Inherent} methods~\cite{chen2019looks,alvarez2018senn,oh2020cbm,kim2018tcav,chattopadhyay2024bootstrapping,ECBM}  learn and deduce concepts alongside the prediction model. These methods necessitate modifications to the models for explanations, posing challenges in scalability to new model architectures and increased computational demands. 

To address these challenges, \emph{post-hoc} methods~\cite{yuksekgonul2022pcbm,pan2021agi,fel2023craft,sundararajan2017ig, bach2015pixel,kindermans2016investigating,rohekar2024causal,xie2022vit-cx,covert2022learning,bennetot2022greybox} deduce concepts from the existing prediction model without additional modifications. Given such advantages, our work focuses on the post-hoc setting. 
These methods are pivotal in image-level explanation for ViTs, providing deeper insights into ViTs' visual data processing. Nevertheless, their focus remains on image-level explanations, overlooking the multi-level structure within ViTs. They also fall short in other desiderata such as faithfulness/stability. 
{Some methods require additional text supervision or human-annotated labels, such as~\cite{yang2023language,ben2024lvlm,menon2022visual,chefer2021transformer,chefer2021generic,ming2022delving,kim2023grounding, losch2019interpretability}. Therefore, these approaches are not applicable to our unsupervised learning setting.}

In contrast, our PACE provides multi-level conceptual explanations that are faithful, stable, sparse, and parsimonious; this is verified by our empirical results in~\secref{sec:experiments}. 


\section{Methodology}
In this section, we formalize the definition of five desiderata for post-hoc conceptual explanations of ViTs and describe our PACE for achieving these desiderata. 


\subsection{Problem Setting and Notations}\label{sec:notations}
\begin{figure}[!t]
        \centering
        \includegraphics[width = 0.44\textwidth]
        {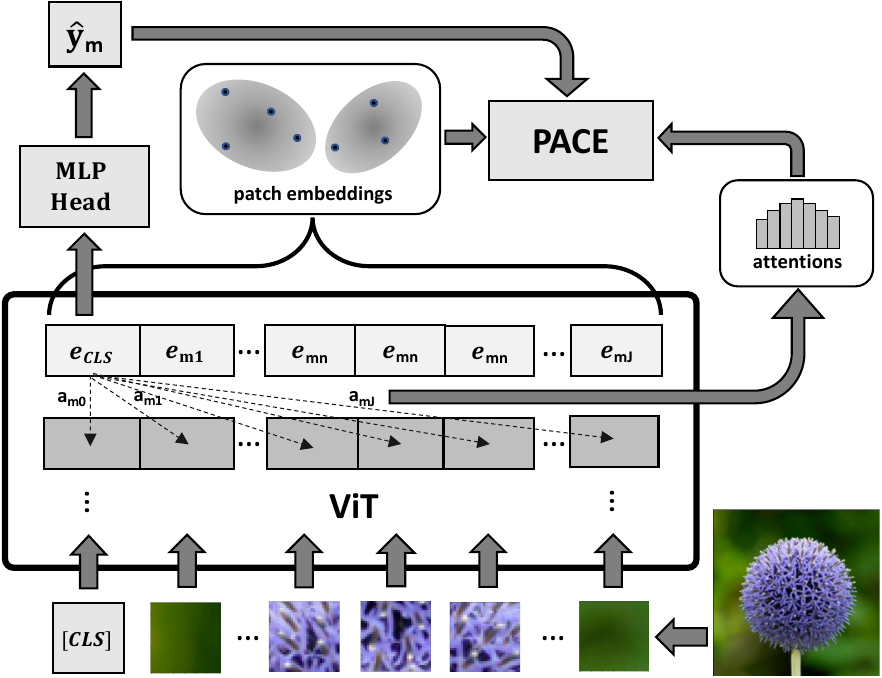}
        \vskip -0.25cm
        \caption[width = 0.8\linewidth]{Overview of PACE framework. {
        PACE utilizes patch embeddings $\e_{m}$, model predictions $\hat\y_m$, and multi-head attentions $\a_m$ as observations to infer hidden parameters.}}
        \label{fig:overview}
        \vskip -0.4cm
\end{figure}

Consider a dataset comprising $M$ images, each dissected into $J$ patches as per the model in~\cite{dosovitskiy2020vit}. We analyze a vision transformer, denoted as $f(\cdot)$, which processes image $m$ (represented by $\I_m$) and yields: (1) the predicted label $\hat \y_m$ with $N$ classes, (2) {patch} embeddings $\e_m \triangleq [\e_{mj}]_{j=1}^{J}$ with $\e_{mj}\in\RB^d$ {($d$ is the hidden dimension)}, and (3) the attention weights $\a_m^{(h)} \triangleq [a_{mj}^{(h)}]_{j=1}^{J}$ ($\a_{m}^{(h)}\in\RB^J$) for each patch relative to the final layer's `CLS' token, where $h$ signifies the attention head $h$. We define the mean attention weight across H heads as $a_{mj}=\frac{1}{H}\sum\nolimits_{h=1}^H a_{mj}^{(h)}$, and consequently $\a_m \triangleq [a_{mj}]_{j=1}^{J}$ (refer to the ViT shown at the bottom of \figref{fig:overview}). 
A typical post-hoc explainer, denoted as $g(\cdot)$, processes the contextual representation $\e_m$, predicted label $\hat \y_m$, and optionally {ViT parameters $\P$}, producing a concept activation $\tha_m\in \RB^{K}$ {($K$ is the number of concepts)}, that is, {$g(\e_m,\hat \y_m,\P)=\tha_m$}. {See~\appref{app:explain_g} for details.} Note that while some methods do not inherently provide explanations with authentic concepts, the explanation activation $\tha_m$ (or its suitably adapted version) can still be interpreted as a quasi-concept vector. 

In contrast to typical post-hoc explainers that only provide image-level explanations $\tha_m$, our PACE provides multi-level conceptual explanations; for an image $m$, PACE provides $K$ \emph{dataset-level} variables $\{\Om_k\}_{k=1}^K=\{\muu_k,\Si_k\}_{k=1}^K$ {($\muu_k\in \RB^{d}$ is the Gaussian mean, and $\Si_k \in \RB^{d\times d}$ the Covariance)}, an \emph{image-level} variable $\tha_m$, and $J$ \emph{patch-level} variables $\ph_m\triangleq\{\ph_{mj}\}_{j=1}^{J}$ {(see details in~\secref{sec:concept_def} and~\secref{sec:experiments})}.  
\subsection{Definition of Trustworthy Conceptual Explanations}
\label{sec:concept_def}
We formally define the five desiderata for trustworthy conceptual explanations for ViTs as follows {(see~\secref{sec:notations})}. 
\begin{definition}[\textbf{Trustworthy Conceptual Explanations}]\label{def:concept} 
Consider a dataset $\DM$ with $M$ images $\I_m$ ($m\in {1,\dots,M}$), each consisting of $J$ patches. For a given number of concepts $K$, a trustworthy conceptual explanation for an image $m$ consists of $K$ \emph{dataset-level} variables $\{\Om_k\}_{k=1}^K=\{\muu_k,\Si_k\}_{k=1}^K$, an \emph{image-level} variable $\tha_m$, and $J$ \emph{patch-level} variables $\{\ph_{mj}\}_{j=1}^{J}$ with the following properties:
\begin{itemize}[noitemsep,leftmargin=17pt]
    \item[(1)] \textbf{Faithfulness}, which implies a strong relation between the concept activation $\tha_m$ and the post-hoc label $\hat y_m$ derived from ViT predictions. In this paper, we measure linear faithfulness score by applying a logistic regression model $LR(\cdot)$, i.e., $\hat y_m=LR(\tha_m) (1\le m \le M)$, and evaluating its accuracy (details in~\secref{sec:experiments}). 
    \item[(2)] \textbf{Stability}, which is the consistency of explanations across different perturbed versions of the same image. For an image $\I_m$ with the inferred $\tha_m$ and its perturbed version $\I'_m$ with inferred $\tha'_m$, stability is quantified by $\frac{\Vert \tha_m-\tha'_m\Vert}{\Vert \tha_m\Vert}$, where $\Vert\cdot\Vert$ denotes the $L_2$ norm. 
    \item[(3)] \textbf{Sparsity}, which involves the concept vector having a sparse representation, measured by the fraction of values nearing zero. Specifically, sparsity is defined as the proportion of $\tha_m$'s entries nearing zero, i.e., $\frac{1}{K}\sum\nolimits_{k=1}^K\mymathbb{1}(\vert \theta_{mk}\vert<\epsilon)$, 
    with a small threshold $\epsilon>0$. 
    \item[(4)] \textbf{Multi-Level Structure}, 
    which means that an ideal explainer should yield 
    $K$ \emph{dataset-level} variables $\{\Om_k\}_{k=1}^K=\{\muu_k,\Si_k\}_{k=1}^K$ representing the mean and covariance of each concept in the dataset, 
    an \emph{image-level} variable $\tha_m\in \RB^K$ for each image $m$, and a \emph{patch-level} variable $\ph_{mj} \in \RB^K$ for each patch $j$ in image $m$. 
    \item[(5)] \textbf{Parsimony}, which involves using the minimal number of concepts $K$ to produce clear and simple explanations for humans. Methods with flexible concept counts could use fewer concepts while maintaining other properties' performance. In contrast, too many concepts usually lead to redundancy in conceptual explanations and a lack of compact representation.
\end{itemize}
\end{definition}

In Definition~\ref{def:concept}, Property (1) ensures the learned concepts convey essential information for predicting image labels from hidden embeddings. Property (2) guarantees robustness and generalization in face of perturbations. Property (3) reflects that each prediction usually only involves a small number of relevant concepts. Property (4) offers diverse and comprehensive multi-level conceptual explanations. Finally, Property (5) facilitates learning concepts efficiently, restricts the number of redundant concepts for meaningful explanations, and reduces humans' cognitive load reading explanations. 
\thmref{thm:MI_bound} in~\secref{sec:theory} provides the theoretical guarantees for PACE in terms of {these properties}.  

\subsection{ProbAbilistic Conceptual Explainers (PACE)}\label{sec:overview}
Drawing inspiration from hierarchical Bayesian deep learning~\cite{BDL,BDLSurvey,NPN,CausalTrans,CounTS,VDI}, we introduce a variational Bayesian framework, dubbed ProbAbilistic Concept Explainers (PACE), for post-hoc conceptual explanation of Vision Transformers (ViTs). To ensure PACE produces trustworthy concepts as defined in Definition~\ref{def:concept}, PACE treats the explained ViT’s patch-level embeddings as observed variable and 
design a hierarchical Bayesian model that generates these embeddings in a top-down manner, from dataset-level concepts through image-level concepts to patch-level embeddings. 

\figref{fig:overview} shows an overview of our PACE, where patch embeddings $\e_m$ and ViT's predicted label $\hat \y_m$ are treated as observable variables. Attention weights $\a_m$ are considered as the {virtual count} for each patch; for example, $a_{mj}=0.2$ means patch $j$ is considered as $0.2J$ patches, where $J$ is the total number of patches in image $m$ {(see details below)}.
PACE models the patch embedding distribution using a mixture of $K$ Gaussians ($K$ concepts), characterized by parameters $ \muu_k\in\RB^d$ and $\Si_k\in\RB^{d\times d}$ $(1\le k \le K)$. 
For image $m$, PACE provides three levels of conceptual explanations: 
(1) $K$ \emph{dataset-level} variables $\{\Om_k\}_{k=1}^K=\{\muu_k,\Si_k\}_{k=1}^K$ representing the mean and covariance of each concept $k$ in the dataset, 
(2) an \emph{image-level} variable $\tha_m\in \RB^K$ for each image $m$, and 
(3) $J$ \emph{patch-level} variable {$\ph_m \triangleq \{\ph_{mj}\}_{j=1}^{J}$} 
for each patch $j$ in image $m$, where $\ph_{mj}\in\RB^{K}$. 

%

\begin{figure}[!t]
        \centering
        \includegraphics[width = 0.4\textwidth]
        {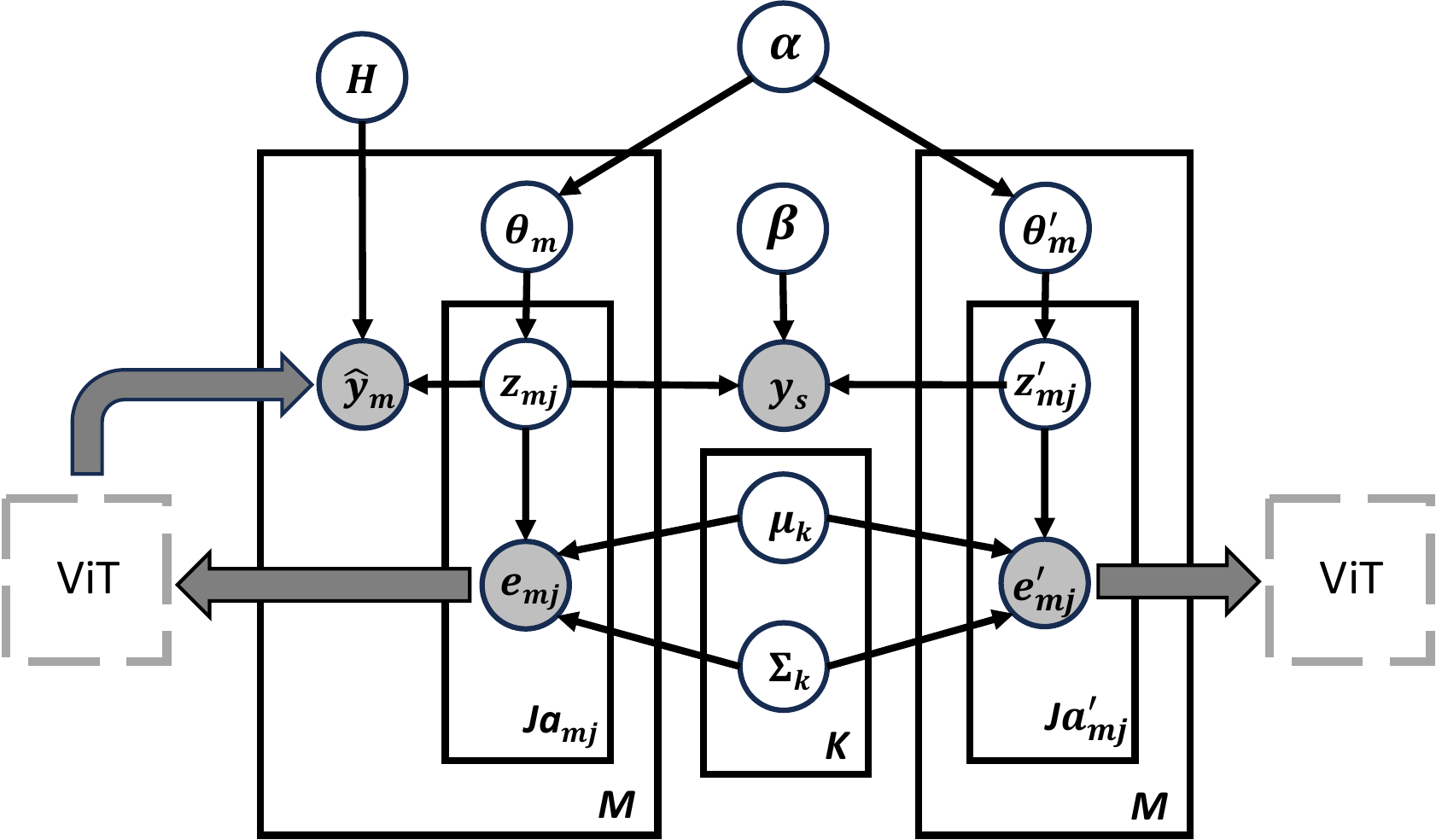}
       \vskip -0.3cm
        \caption{Graphical model of our PACE. We sample each original patch embedding $\e_{mj}$ for $J\cdot a_{mj}$ times, and each perturbed patch embedding $\e'_{mj}$ for $J\cdot a'_{mj}$ times (ViT is shared for both).}
        \label{fig:pgm}
        \vskip -0.4cm
\end{figure}

\textbf{Generative Process.}
Below we describe the generative process of PACE (\figref{fig:pgm} shows the corresponding PGM): 

\begin{itemize}
\item Draw the image-level concept distribution vector $\tha_m\sim \text{Dirichlet}(\alp)$ for either the original image $\I_m$ or the perturbed image $\I'_m$.
\item For each patch $j$ in either $\I_m$ or $\I'_m$ $(1\leq j \leq J)$:
\begin{itemize}
\item Draw the patch-level one-hot concept index
$\z_{mj}\sim \text{Categorical}(\tha_m)$.
\item Given the ViT attentions $a_{mj}$, for $J\cdot a_{mj}$ times, 
\begin{itemize}
    \item Draw patch $j$'s embedding, i.e., $\e_{mj}$, from concept $\z_{mj}$'s Gaussian component $\e_{mj}\sim\mathcal{N}(\muu_{z_{mj}},\Si_{z_{mj}})$.
\end{itemize}

\end{itemize}
\item Draw the predicted label $\hat \y_m \sim \text{GLM}(\bar\z_m, \mH)$. 
\item  For each pair of images $\I_m$ and $\I'_{m}$, 
draw a binary variable $y_s \sim p(y_s|\bar\z_m,\bar\z'_{m},\bet) $, which indicates whether $\I_{m}$ and $\I'_{m}$ come from the same image.

\end{itemize}
Here $\alp\in \RB^K$ is the parameter for the Dirichlet distribution $\text{Dirichlet}(\cdot)$, and we define
\begin{align} \label{z_avg_def}
\bar\z_m = \nicefrac{1}{J}\sum\nolimits_{j=1}^{J} \z_{mj}.
\end{align}
$\text{GLM}(\cdot)$ denotes a categorical distribution from a generalized linear model (GLM), given by
\begin{align}\label{eq:prob_class}
    p(\hat\y_m|\mH,\bar \z_m) = \prod\nolimits_{n=1}^{N}[\frac{\exp(\et_n^T \bar\z_{m})}{\sum\nolimits_{n'}\exp(\et_{n'}^T \bar\z_{m})}]^{\hat y_{mn}},
\end{align}
where $N$ is the number of classes, and $\mH=[\et_{1},...,\et_N]$  are the learnable parameters ($\mH \in \RB^{K\times N}$).
The function $p(y_s|\bar\z_{m},\bar\z'_m,\bet)$ defines a distribution over whether $\I'_{m}$ is the perturbation of $\I_m$, where $y_s$ is a binary label. Let $\FM = \{1,2,..,M\} \backslash \{m\}$, and we have $p(y_s|\bar\z_{m},\bar\z'_m,\bet)$ as
\begin{align}\label{eq:prob_indicator}
    p(y_s=1|\bar\z_{1:M}, \bar\z'_m,\bet) 
    =\frac{\exp(\bet^T (\bar\z_{m}\circ\bar\z'_{m}))}{\sum\nolimits_{f\in \FM} \exp(\bet^T (\bar\z_{m}\circ\bar\z_{f}))},
\end{align}
where $\bet\in \mathbb{R}^K$ is a learnable parameter, and $\circ$ denotes the element-wise product.

Given this generative process, learning the latent concept structures in ViT \emph{across the dataset} involves learning the dataset-level parameters $\{\muu_k,\Si_k\}_{k=1}^K$ for the K concepts. Similarly, explaining ViT \emph{for each image} is equivalent to inferring the distributions of the image-level and patch-level latent variables $\tha_m$ and $\{\z_{mj}\}_{j=1}^J$, respectively.




\subsection{Inferring Conceptual Explanations using PACE} \label{sec:inference}

We begin by detailing the inference of image-level and patch-level explanations (i.e., $\tha_m$ and {$\{\z_{mj}\}_{j=1}^J$}) given the dataset-level concept parameters $\{\muu_k,\Si_k\}_{k=1}^K$. We then discuss learning $\{\muu_k,\Si_k\}_{k=1}^K$ later in \secref{sec:learning}. 

\textbf{Inferring Patch-Level and Image-Level Concepts.} 
Given the dataset-level concept parameters $\{\muu_k,\Si_k\}_{k=1}^K$, the  patch embeddings $\e_m \triangleq [\e_{mj}]_{j=1}^{J}$, and the associated attention weights $\a_m \triangleq [a_{mj}]_{j=1}^{J}$, as well as the predicted label $\hat \y_m$ produced by the ViT, for each image $\I_m$, PACE infers the posterior distribution of the image-level concept explanation $\tha_m$, i.e., $p(\tha_m | \e_m,\a_m,\{\muu_k,\Si_k\}_{k=1}^K, \hat \y_m)$, and the posterior distribution of the patch-level concept explanation $\z_{mj}$, i.e., $p(\z_{mj} | \e_m,\a_m,\{\muu_k,\Si_k\}_{k=1}^K,\hat \y_m)$. {\figref{fig:overview} describes the inference process of PACE}. 

\textbf{Variational Distributions.} 
The aforementioned posterior distributions are intractable; hence, we employ variational inference~\citep{VI,blei2003latent,chang2009rlda}, using variational distributions $q(\tha_m | \gamm_m)$ and $q(\z_{mj} | \ph_{mj})$ to approximate them. This results in the following joint variational distribution:
\begin{align}
&q(\tha_m, \{\z_{mj}\}_{j=1}^{J}|\gamm_m, \{\ph_{mj}\}_{j=1}^{J}) \nonumber\\
=~& q(\tha_m | \gamm_m) \cdot \prod\nolimits_{j=1}^{J}q(\z_{mj} | \ph_{mj}),\label{eq:joint_vi}
\end{align}
where the variational parameters $\gamm_m\in \RB^K$ and $\ph_{mj}  \in \RB^K$ 
are estimated by maximizing \eqnref{eq:full_elbo}  (more details below). 

\textbf{Objective Function.} 
In line with the generative process outlined in~\secref{sec:overview}, for each image $m$ sampled from the dataset, the optimal variational distributions are found by maximizing the following evidence lower bound (ELBO):
%
\begingroup\makeatletter\def\f@size{9}\check@mathfonts
\begin{align}\label{eq:full_elbo}
    L(\e_{mj}&, \gamm_m, \ph_m,\ph'_{m},\hat \y_{m}, y_s; 
    \alp, \{\muu_k,\Si_k\}_{k=1}^K, \mH,\bet) \nonumber \\
    =~&L_e + L_f + L_s, \\ \label{eq:full_elbo_expand}
    L_e=~& L(\e_{mj},\gamm_m, \ph_{m}; \alp, \{\muu_k,\Si_k\}_{k=1}^K),
    \nonumber \\
    L_f=~& L(\hat \y_{m}, \ph_m; \mH),\nonumber \\
    L_s=~& L(y_s, \ph_{m},\ph'_{m};\bet).
\end{align}
\endgroup
%
This equation can be derived using log likelihood factorization {of the variables in~\figref{fig:pgm}} (details provided in~\appref{app:derivation}). 
Below we describe each term's intuition:
\begin{enumerate}[nosep]
\item $L_e = L(\e_{mj}, \gamm_m, \ph_m; \alp, \{\muu_k,\Si_k\}_{k=1}^K)$ is the expected log likelihood of the joint distribution of patch embeddings $\e_{mj}$ and the variational parameters $\gamm_m,\ph_m$. 
This term models the generation of patch embeddings $\e_{mj}$ in the ViT. 
\item 
$L_f = L(\hat \y_{m}, \ph_m; \mH)$ is the expected log likelihood of the predicted label $\hat \y_m$ given explanation $\ph_m$. This term reflects the \textbf{\emph{faithfulness}} property in~\defref{def:concept}.
\item 
$L_s = L( y_s, \ph_{m}, \ph'_{m};\bet)$ is the expected log likelihood of the binary label $y_s$, 
which indicates whether image $\I_{m}$ (with its inferred concepts $\ph_m$) and $\I'_{m}$ (with its inferred concepts $\ph'_m$) comes from the same image. 
This term reflects the \textbf{\emph{stability}} property in~\defref{def:concept}.
\end{enumerate}


\textbf{Computing $L_e$.} We compute $L_e$ as: 
\begingroup\makeatletter\def\f@size{7}\check@mathfonts
\begin{align}
    L_e = ~&\EB_q[\log p(\e_{mj},\gamm_m, \ph_m|\alp,\{\muu_k,\Si_k\}_{k=1}^K)]  \nonumber\\
    =~&\sum\nolimits_{{k}} \phi_{mjk}a_{mj}\log\NM(\e_{mj}|\muu_k,\Si_k) 
    +\EB_q[\log p(\gamm_m, \ph_m\vert\alp)]
    \nonumber\\
    =~&\sum\nolimits_{{k}}\phi_{mjk} a_{mj} \{-\frac{1}{2}(\e_{mj}-\muu_k)^T\Si_k^{-1}(\e_{mj}-\muu_k) \nonumber\\
   & -\log[(2\pi)^{d/2} \vert \Si_k\vert^{1/2}]\} 
   +\EB_q[\log p(\gamm_m, \ph_m\vert\alp)], \label{eq:elbo}
\end{align}
\endgroup
where the expectation is over the joint variational distribution in~\eqnref{eq:joint_vi}. $d$ is the dimension of the embedding $\e_{mj}$.

\textbf{Computing $L_f$.} We compute $L_f$ according to~\eqnref{eq:prob_class}: 
\begingroup\makeatletter\def\f@size{7.5}\check@mathfonts
\begin{align}\label{eq:faithful}
    L_f =& \mathbb{E}_{q}[\log p(\hat\y_m|\bar\z_m,\mH)] \nonumber \\
     =& \sum\nolimits_{n=1}^{N} \hat y_{mn} (\et^T_{n}  \bar \ph_m) 
     -  \EB_{q} [\log (\sum\nolimits_{n=1}^{N} \exp (\et^T_{n} \bar\z_m))] \nonumber\\
     \approx & \sum\nolimits_{n=1}^{N} \hat y_{mn} (\et^T_{n}  \bar \ph_m) 
     -   \log (\sum\nolimits_{n=1}^{N} \exp (\et^T_{n} \bar\ph_m)),     
\end{align}
\endgroup
where $N$ is the number of classes for classification, $n$  the class index. We approximate $\bar\z_m$ by taking the average of $\ph_m$: $\bar \z_m \approx \bar \ph_m = \nicefrac{1}{J}\sum\nolimits_{j=1}^{J} {\ph}_{mj}.$ See~\appref{app:derivation} for details on how to approximate $\bar \z_m$. {\eqnref{eq:faithful} implies that maximizing the log likelihood of the predicted class $\hat\y_m$ encourages a correlation between $\hat\y_m$ and the inferred patch-level concepts $\ph_m$, thereby enhancing PACE's \emph{faithfulness.}}

\textbf{Computing $L_s$.} 
Inspired by contrastive learning~\cite{chen2020simclr}, for each original image $\I_m$, we first generate its perturbed image $\I'_m$. 
Then, with their associated patch-level concepts $\bar \z_{1:M}$ and $\bar \z'_m$ from~\eqnref{z_avg_def}, the \emph{stability} term $L_s$ is defined as the expected likelihood of the {binary label} $y_s$ in~\eqnref{eq:prob_indicator}.  
Let $\FM = \{1,..,M\} \backslash \{m\}$. We compute $L_s$ as:
\begingroup\makeatletter\def\f@size{7.5}\check@mathfonts
\begin{align}\label{eq:stability}
 L_s =~& \EB_{q}[\log p(y_s=1\vert \bar \z_{1:M},\bar \z'_m, \bet)] 
 \nonumber\\
    =~&\bet^T (\bar\z_{m}\circ\bar\z'_{m})
     -  \EB_{q} [\log (\sum\nolimits_{f\in \FM} \exp(\bet^T (\bar\z_{m}\circ\bar\z_{f})) )] \nonumber\\
     \approx~& \bet^T (\bar\ph_{m}\circ\bar\ph'_{m})
     - \log (\sum\nolimits_{f\in \FM} \exp(\bet^T (\bar\ph_{m}\circ\bar\ph_{f})) ),
\end{align}
\endgroup
where $\circ$ is the element-wise product. {\eqnref{eq:stability} indicates that maximizing the log likelihood of $y_s$ encourages the inferred patch-level concepts from the original and perturbed patches ($\ph_m$ from $\I_m$ and $\ph'_m$ from $\I'_m$) to be similar, thus enhancing PACE's \emph{stability} against perturbations.}
See detailed derivations of $L_e$, $L_f$, and $L_s$ in~\appref{app:derivation}.

\begin{algorithm}[!t]
 \caption{Learning and Inference of PACE}\label{alg:vace}
 \textbf{Input:} Initialized $\alp, \bet, \mH, \{\gamm_m\}_{m=1}^M$, $\{\ph_m\}_{m=1}^M$, $\{\Om_k\}_{k=1}^K$, images $\{\I_m\}_{m=1}^M$, perturbed images $\{\I'_m\}_{m=1}^M$, predicted labels $\{\hat \y_m\}_{m=1}^M$, and number of epochs~T.
 
 \textbf{for} {$t=1:T$} \textbf{do}{
 
 \quad\textbf{for} {$m=1:M$} \textbf{do}{
 
 \quad\quad Update $\ph_m$ and $\gamm_m$ using \eqnref{eq:update_phi} and \eqnref{eq:update_gamma}, respectively.
 
 }
 \quad Update $\{\Om_k\}_{k=1}^K$ using \eqnref{eq:update_mu} and \eqnref{eq:update_sigma}.

\textbf{Output:} $\{\Om_k\}_{k=1}^K$ as dataset-level, $q(\tha_m | \gamm_m)$ as image-level, and $q(\z_m|\ph_m)$ as patch-level concept explanations.
 }
\end{algorithm}
\textbf{Update Rules for $\phi_{mjk}$ and $\gamma_{mk}$.} 
Inferring the conceptual explanations using PACE involves learning the variational parameters, $\phi_{mjk}$ and $\gamma_{mk}$, in \eqnref{eq:joint_vi}. 
This is done by 
iteratively updating $\phi_{mjk}$ and $\gamma_{mk}$ 
to maximize \eqnref{eq:full_elbo}. 

Specifically, taking the derivative of the ELBO in \eqnref{eq:full_elbo} with respect to $\phi_{mjk}$ (see~\appref{app:inference} for details) and setting it to zero, we obtain the update rule for $\phi_{mjk}$:
\begingroup\makeatletter\def\f@size{8.5}\check@mathfonts
\begin{align}\label{eq:update_phi}
    \phi_{mjk} \propto ~& \frac{1}{\vert \Si_k\vert^{1/2}} 
      \exp[\Psi(\gamma_{mk})-\Psi(\sum\nolimits_{k'=1}^K \gamma_{k'}) \nonumber\\
     - &\frac{1}{2}a_{mj}(\e_{mj}-\muu_k)^T\Si_k^{-1}(\e_{mj}-\muu_k)\nonumber\\ 
     + & \frac{1}{J} [(\sum\nolimits_{n=1}^{N} \hat{y}_{mn} \et_n 
    -  \frac{\sum\nolimits_{n=1} ^{N} \exp(\et^T_{n} \bar\ph_{m}){\et_n}} {\sum\nolimits_{n=1}^{N} \exp (\et^T_{n} \bar\ph_m)})_k\nonumber\\
     + &\bet^T \bar \ph'_m  - \frac{\sum\nolimits_{f\in \FM}\exp (\bet^T  (\bar \ph_m \circ \bar\ph_f)  ) (\bet^T \bar\ph_f)} {\sum\nolimits_{f\in \FM}\exp (\bet^T (\bar\ph_m \circ \bar\ph_f ))}) ],
\end{align}
\endgroup
with the normalization constraint $\sum\nolimits_{k=1}^K \phi_{mjk}=1$. Here $\Psi(\cdot)$ is the digamma function (the first derivative of the logarithm of the Gamma function $\Gamma(z) = \int_0^\infty t^{z-1} e^{-t} dt$).

Similarly, 
the update rule for $\gamma_{mk}$ is:
\begin{align}
    \gamma_{mk} = \alpha_k + \sum\nolimits_{j=1}^{J}  \phi_{mjk}a_{mj}.\label{eq:update_gamma}
\end{align}

In summary, the inference algorithm alternates between updating $\phi_{mjk}$ for all $(m,j,k)$ tuples and updating $\gamm_{mk}$ for all $(m,k)$ tuples until convergence. 

\textbf{Image- and Patch-Level Explanations: $\tha_m$ and $\ph_{mj}$.} 
We then use $\gamm_m=\{\gamma_{mk}\}_{k=1}^K$ with $q(\tha_m | \gamm_m)$ to obtain the \emph{image-level} conceptual explanation $\tha_m$ and use $\ph_{mj}=\{\phi_{mjk}\}_{k=1}^K$ as the \emph{patch-level} explanation. 

\subsection{Learning of PACE}\label{sec:learning}

\textbf{Learning Dataset-Level Explanations: $\{\muu_k,\Si_k\}_{k=1}^K$.} 
The inference algorithm in~\secref{sec:inference} assumes the availability of the dataset-level concept parameters $\{\muu_k,\Si_k\}_{k=1}^K$. To learn these parameters, one needs to iterate between (1) inferring image-level and patch-level variational parameters $\gamm_m$ and $\ph_{mj}$ in \secref{sec:inference}, respectively, and (2) learning dataset-level concept parameters $\{\muu_k,\Si_k\}_{k=1}^K$. \algref{alg:vace} summarizes the learning of PACE.

\textbf{Update Rules for $\muu_k$ and $\Si_k$.} Similar to~\secref{sec:inference}, we expand the ELBO in~\eqnref{eq:elbo} (see~\appref{app:learning} for details) and set its derivative with respect to $\muu_k$ and $\Si_k$ to zero,
yielding the update rule for learning $\muu_k$ and $\Si_k$:
\begin{align}
    &\muu_k =  \frac{\sum\nolimits_{m,j}{\phi_{mjk}a_{mj} \e_{mj}}}{\sum\nolimits_{m,j} \phi_{mjk}a_{mj}}, \label{eq:update_mu} \\
    &\Si_k = \frac{\sum\nolimits_{m,j}\phi_{mjk}a_{mj} (\e_{mj}-\muu_k)(\e_{mj}-\muu_k)^T}{\sum\nolimits_{m,j} \phi_{mjk}a_{mj} }. \label{eq:update_sigma}
\end{align}

\subsection{Summary of Learning and Inference of PACE}

In summary, PACE is a variational Bayesian framework that consists of (1) the \textbf{learning stage} to train on the training set, and (2) the \textbf{inference stage} to explain on the test set. 

For example, given a finetuned ViT classifier on a dataset, PACE explains it by (1) training the global parameters, i.e., the \emph{dataset-level} concept centers $\muu_k$ and covariance matrices $\Si_k$ (where $k=1, \dots, K$) as dataset-level explanations, on the training set (these are called \emph{global} parameters because they are shared across all data points, e.g. images), and (2) inferring the \emph{local} parameters, i.e., the \emph{image-level} concepts (explanations) $\tha_m$ and \emph{patch-level} concepts (explanations) $\ph_m$, on the test set (these are called \emph{local} parameters because each image has its own $\tha_m$ and $\ph_m$). 

Below, we discuss the learning and inference processes, respectively. 

\textbf{The Learning Stage.} In~\secref{sec:learning}, we describe the process of learning the \emph{global parameters} $\muu_k$ and $\Si_k$ (where $k=1, \dots, K$). As described in~\algref{alg:vace}, in each epoch $t$ ($t=1,\dots,T$):
\begin{itemize}
    \item[(1)] PACE first infers the \emph{local} parameters $\gamm_m$ and $\ph_{mj}$ for each document $m$ using~\eqnref{eq:update_phi} and~\eqnref{eq:update_gamma}; 
    \item[(2)] PACE then updates the \emph{global} parameters $\muu_k$ and $\Si_k$ (where $k=1, \dots, K$) for the entire dataset using~\eqnref{eq:update_mu} and~\eqnref{eq:update_sigma}.
\end{itemize}
The learning stage concludes at the $T^{th}$ epoch. Note that the process above is iterative; it alternates between (1) updating the \emph{local} parameters and (2) updating the \emph{global} parameters.

\textbf{The Inference Stage.} In~\secref{sec:inference}, we describe the process of inferring the \emph{local} parameters $\tha_m$ and $\ph_{mj}$ after the PACE learns the global parameters in the learning stage. Specifically, given the \emph{global} parameters $\muu_k$ and $\Si_k$ (where $k=1, \dots, K$), 
\begin{itemize}
    \item[(1)] PACE initializes \emph{local} parameters $\gamm_m$ and $\ph_{mj}$;
    \item[(2)] Given the current $\gamm_m$ and $\ph_{mj}$, PACE updates the \emph{local parameters} $\gamm_m$ using~\eqnref{eq:update_phi};
    \item[(3)] Given the current $\gamm_m$ and $\ph_{mj}$, PACE updates the \emph{local parameters} $\ph_{mj}$ using~\eqnref{eq:update_gamma};
    \item[(4)] PACE repeats (2) and (3) until $\gamm_m$ and $\ph_{mj}$ converge; 
    \item[(5)] PACE then infers the image-level concept $\tha_m$ using the learned variational distribution $q(\tha_m|\gamm_m)$, which is a Dirichlet distribution. One can (roughly) think of $\tha_m$ as a normalized version of $\gamm_m$. 
\end{itemize}





\subsection{Discussion and Theoretical Analysis}\label{sec:theory}
Our PACE addresses all five desiderata in \defref{def:concept}:
\begin{itemize}[nosep]
\item \textbf{Faithfulness} is encouraged by 
maximizing the prediction $\hat \y_{m}$'s likelihood, i.e., $L_f$ in~\eqnref{eq:faithful}. 
\item \textbf{Stability} {against perturbations is enhanced by maximizing the {binary label} $y_s$'s likelihood $L_s$ in~\eqnref{eq:stability}}.
\item \textbf{Sparsity} is encouraged by the Dirichlet prior $p(\tha_m|\alp)$ that regularizes the inference of $\tha_m$. 
\item \textbf{Multi-Level Structure} is intrinsically supported by our multi-level generative process in \secref{sec:overview}. 
\item \textbf{Parsimony} is ensured by the flexibility in choosing the number of concepts $K$ in PACE (see~\appref{app:implement}).
\end{itemize}

Theorem \ref{thm:MI_bound} below further demonstrates that PACE's inferred image-level and patch-level explanations, $\tha_m$ and $\{\ph_{mj}\}_{j=1}^{J}$, align with the properties in~\defref{def:concept}. 
\begin{theorem}[\textbf{Mutual Information Maximization}]
\label{thm:MI_bound}
The ELBO 
    in~\eqnref{eq:full_elbo} is upper-bounded by the sum of (1) mutual information between contextual embeddings $\e_m$ and multi-level explanation $\tha_m,\{\ph_{mj}\}_{j=1}^{J}$ in~\defref{def:concept}, (2) mutual information between the predicted label $\hat \y_m$ and patch-level concept $\ph_m$, and (3) mutual information between the patch-level original concept $\ph_m$ and perturbed concept $\ph'_m$. Formally, with approximate posteriors $q(\tha_m | \gamm_m)$ and $q(\z_{mj} | \ph_{mj})$, we have
\begingroup\makeatletter\def\f@size{9.5}\check@mathfonts
\begin{align}\label{eq:MI_bound}
&L(\e_{mj}, \gamm_m, \ph_m,\ph'_{m},\hat \y_{m}, y_s)\nonumber\\
     \leq ~& {I}(\e_m; \tha_m,\ph_m) + I(\hat \y_m;\ph_m) 
     + I(\ph_m;\ph'_m) 
     +  C,
\end{align}
\endgroup
where the $C$ is a constant.
\end{theorem}


The proof of~\thmref{thm:MI_bound} is provided in~\appref{app:proof}. 
\thmref{thm:MI_bound} implies that maximizing the ELBO in~\eqnref{eq:full_elbo} is equivalent to maximizing the sum of (1) mutual information between the contextual embeddings $\e_m$ and \emph{\textbf{multi-level}} conceptual explanations defined in~\defref{def:concept}, thereby ensuring the generated explanations are informative, (2) mutual information between the ViT prediction $\hat \y_m$ and patch-level concept $\ph_m$, thereby enhancing PACE's \emph{\textbf{faithfulness}}, and (3) mutual information between the patch-level original $\ph_m$ and perturbed $\ph'_m$, thereby enhancing PACE's \emph{\textbf{stability}} against perturbations.


\section{Experiments}\label{sec:experiments}
In this section, we compare PACE with existing methods on one synthetic dataset and three real-world datasets. 

\subsection{Datasets}\label{sec:datasets}

\begin{figure}[t]
  \centering
  \includegraphics[width=0.8\linewidth]{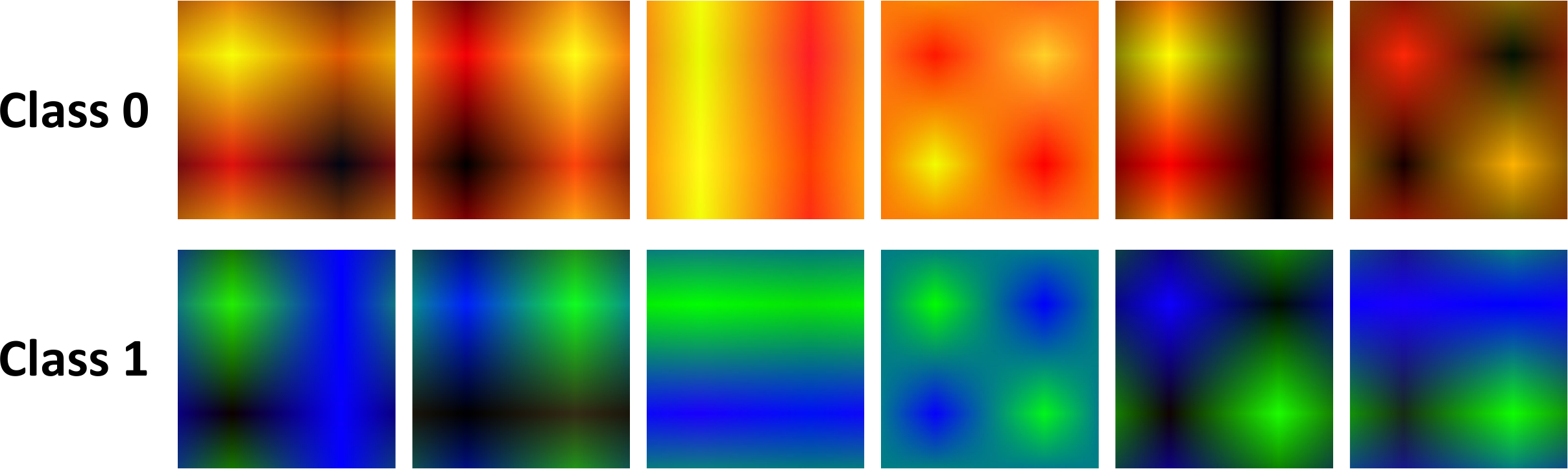}
  \caption{Example images from our \emph{Color} dataset.} 
  \label{fig:color_samples}
  \vskip -0.3cm
\end{figure}

We constructed \emph{Color} as a synthetic dataset with clear definition of 4 concepts (\red{red}/\yellow{yellow}/\green{green}/\blue{blue}). It contains two image classes: \emph{Class 0} (images with \red{red}/\yellow{yellow} colors) and \emph{Class 1} (\green{green}/\blue{blue} colors), both against a \emph{black} background (see example images in~\figref{fig:color_samples} and more details in~\appref{app:implement}). 
We use three real-world datasets, \emph{Oxford 102 Flower ({Flower})}~\cite{nilsback2008oxford102flower}, \emph{Stanford Cars ({Cars})}~\cite{krause2013stanfordcars}, and \emph{CUB-200-2011 (CUB)}~\cite{CUB} (for dataset statistics, see~\appref{app:implement}). 
For real-world datasets, we follow the preprocessing steps from~\cite{dosovitskiy2020vit} and use the same train-test split. For the \emph{Color} dataset, we adopt an 8:2 train/test split among 2,000 images (1,000 per class). 

\begin{figure*}[!t]
        \centering
        \includegraphics[width = 0.95\textwidth]{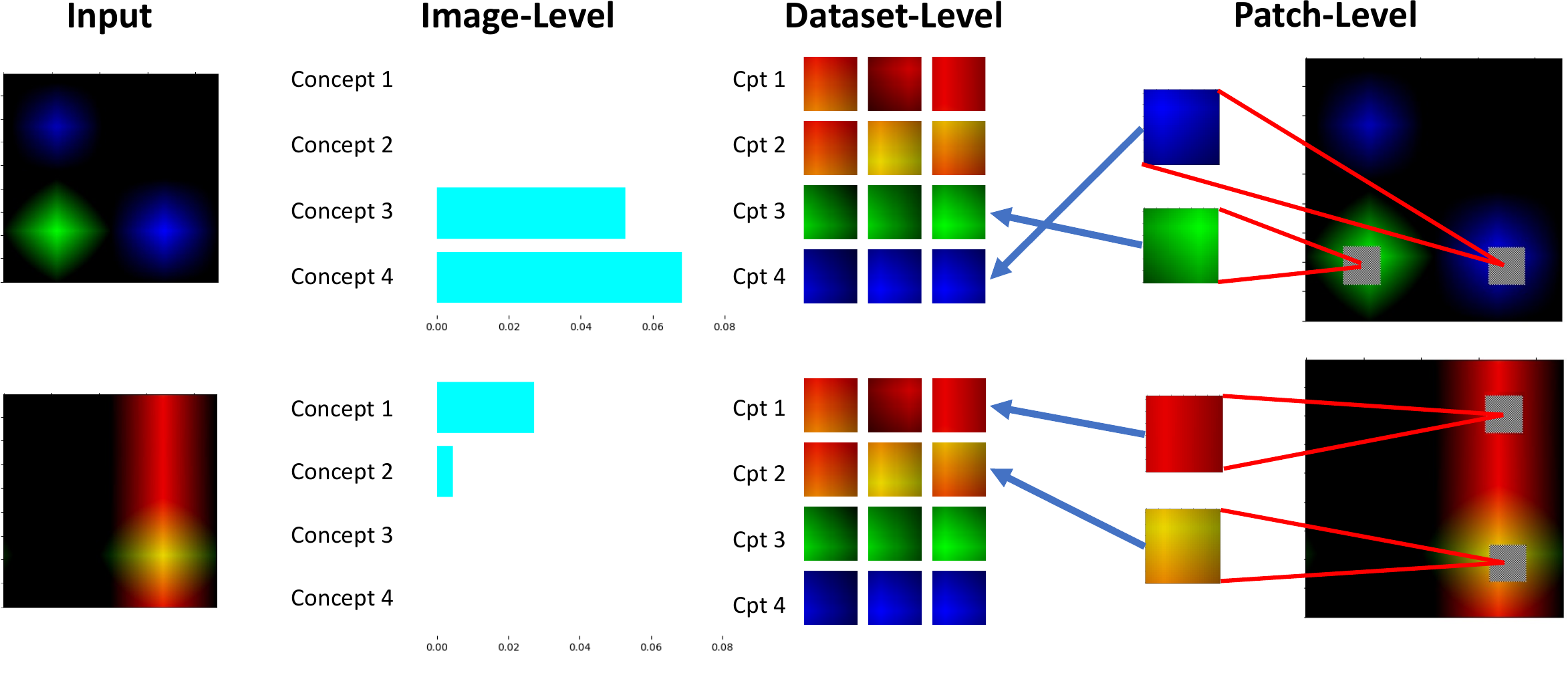} 
        \vskip -0.6cm
        \caption[width = 0.8\linewidth]{PACE's three-level conceptual explanations on the \emph{Color} dataset. \textbf{Dataset-Level}: PACE's top 4 dataset-level concepts; for each concept $k$, we plot the top 3 patches with $\e_{mj}$ closest to $\muu_k$. \textbf{Image-Level}: 
        Given an input image $m$, we show PACE's generated image-level explanation $\theta_m$ for the 4 selected concepts. For example, for the top-left input image, PACE's generated image-level explanation $\theta_m$ indicates a strong association with Concept $3$ (\textcolor{green}{green}) and Concept $4$ (\textcolor{blue}{blue}). \textbf{Patch-Level}: 
        Given an input image $m$, PACE's $\ph_{mj}$ identifies the top concepts for the selected patches. For example, for the top-left input image, the \blue{blue} patch is associated with Concept $4$ (containing similar \textcolor{blue}{blue} patches across the dataset) while the \textcolor{green}{green} patch is linked to Concept $3$ (containing similar \textcolor{green}{green} patches across the dataset). }
        \label{fig:color_qualitative}
        \vskip -0.4cm
\end{figure*}

\begin{table*}[!t]
\caption{Results for the five desiderata in~\defref{def:concept} for different methods on four datasets. `All' denotes results on all four datasets. We mark the best results with \textbf{bold face} and the second best results with \underline{underline}. {$\uparrow/\downarrow$ indicates higher/lower is better, respectively.}}\label{tab:real_world_compare}
\vskip  0.1 cm
\centering
\resizebox{0.99\textwidth}{!}{%
\begin{tabular}{lccccccccccccccc}
\toprule
Desiderata & \multicolumn{4}{c}{(Linear) Faithfulness~$\boldsymbol{\uparrow}$} & \multicolumn{4}{c}{Stability~$\boldsymbol{\downarrow}$} & \multicolumn{4}{c}{Sparsity~$\boldsymbol{\uparrow}$} & \multicolumn{1}{c}{Multi-Level} & \multicolumn{1}{c}{Parsimony~$\boldsymbol{\downarrow}$} \\
\cmidrule(r){1-1} \cmidrule(r){2-5} \cmidrule(r){6-9} \cmidrule(r){10-13} \cmidrule(r){14-14} \cmidrule(r){15-15}
Datasets & \emph{Color} & \emph{Flower} & \emph{Cars} & \emph{CUB} & \emph{Color}& \emph{Flower} & \emph{Cars} & \emph{CUB} & \emph{Color} & \emph{Flower} & \emph{Cars} & \emph{CUB} & All & All\\
\midrule
SHAP & 0.47 &0.52& 0.44 & 0.34& 4.39& 0.92 &1.21 &1.55 &0.54 & 0.12 & 0.13 & 0.11 & No& 768 \\
LIME &0.54&0.06 & 0.02& 0.03& 1.50 & 1.54& 1.45& 1.80& \underline{0.59}&\bf{0.54} &\bf{0.52} &\underline{0.55} & No&768 \\
Saliency & \bf{1.00} & \underline{0.57} & \bf{0.50}&\underline{0.49} &\underline{0.35}& \underline{0.47}& \underline{0.43}&\underline{0.48} & 0.01&0.00&0.00 &0.00 & No& 768 \\
AGI & \bf{1.00} & {0.54}& 0.34& \underline{0.49}&1.40 &1.21 &1.83 & 2.53&0.01 &0.07 & 0.04& 0.03& No&768  \\
CRAFT & 0.59& 0.01&0.01 & 0.00&4.49&{0.52} & 1.76&0.64 & 0.26&0.29 & 0.11&0.25 &{Partial} &\bf{100}  \\
\midrule
{PACE}  & \bf{1.00} &\bf{0.80}& \bf{0.50}&\bf{0.56}&\bf{0.20} &\bf{0.12} &\bf{0.05} & \bf{0.05}& \bf{0.97}&\underline{0.48}&\underline{0.49} &\bf{0.63} &\textbf{{Full}} & \bf{100} \\
\bottomrule
\end{tabular}%
}
\vskip -0.5cm
\end{table*}

\subsection{Baselines} 
We compare PACE with state-of-the-art methods, including: 
\begin{itemize}[nosep,leftmargin=18pt]
\item \textbf{SHAP}~\cite{lundberg2017shap} is an explanation method that assigns importance scores to input features using Shapley values. 
\item \textbf{LIME}~\cite{ribeiro2016lime} explains the model by approximates it with a local surrogate model via data perturbation. 
\item \textbf{SALIENCY}~\cite{simonyan2013saliency} uses the saliency map of an image to explain the model prediction. 
\item \textbf{AGI}~\cite{pan2021agi} produces explanations via adversarial gradient integration. 
\item \textbf{CRAFT}~\cite{fel2023craft} employs recursive low-rank matrix factorization to obtain concepts from intermediate layers. 
\end{itemize}

\subsection{Evaluation Metrics}\label{sec:metrics}
With ViT-Base~\citep{dosovitskiy2020vit} as the prediction model, we evaluate different methods against the five desiderata {defined} in \defref{def:concept}: 
\begin{itemize}
    \item \textbf{(Linear) Faithfulness:} 
    We fit a logistic regression (LR) model $\hat \y_m = LR(\tha_m)$ to each dataset's training set, with $\hat \y_m$ as the prediction of the ViT, and test the model's accuracy on the test set. Higher accuracy indicates stronger (linear) faithfulness. Note that one can fit more complex models (e.g., nonlinear models such as neural networks) to evaluate nonlinear faithfulness; for simplicity, we focus on linear faithfulness. 
        \textbf{Stability:} 
    For each input image $\I_m$ in the test set, we generate an augmented version $\I'_m$ (following~\citet{chen2020simclr}), and compute the normalized difference between inferred concepts $\tha_m$ and $\tha'_m$, i.e., $\frac{\Vert \tha_m-\tha'_m\Vert}{\Vert \tha_m\Vert}$. Lower values indicate stronger stability. 
    
    \item \textbf{Sparsity:} 
    We compute sparsity (with the threshold $\epsilon=0.1/K$) as $\frac{1}{K}\sum\nolimits_{k=1}^K\mymathbb{1}(\vert \theta_{mk}\vert<\epsilon)$, where K is the number of concepts. 
    For many concept-based explanation methods, including ours, the inferred activation typically normalizes to sum up to 1. If this is not the case, we normalize the explanation activation before calculating the sparsity score, ensuring a fair comparison.

    \item \textbf{Multi-Level Structure:} As highlighted in~\secref{sec:intro}, baseline models do not account for {dataset-level and/or patch-level concepts, thereby possess \emph{No} or \emph{Partial} multi-level structure}. In contrast, PACE is specifically designed to offer {\emph{Full}} conceptual explanations at three levels: dataset, image, and patch. We will demonstrate in~\secref{sec:qualitative} that {modeling embeddings' distribution} is instrumental in bridging three levels of {ViT concepts}.
    \item \textbf{Parsimony:} For the conceptual explanation methods PACE and CRAFT~\cite{fel2023craft}, we set the number of concepts $K=100$, to facilitate a fair comparison. 
    Note that other baseline models' number of concept $K$ is constrained to the hidden dimension of ViT embeddings, i.e., $K=768$; they therefore fall short in parsimony.
\end{itemize}
{For details and the three other desiderata, see~\appref{app:implement}}.

\begin{figure*}[!t]
        \centering
        \includegraphics[width = 1.0\textwidth]{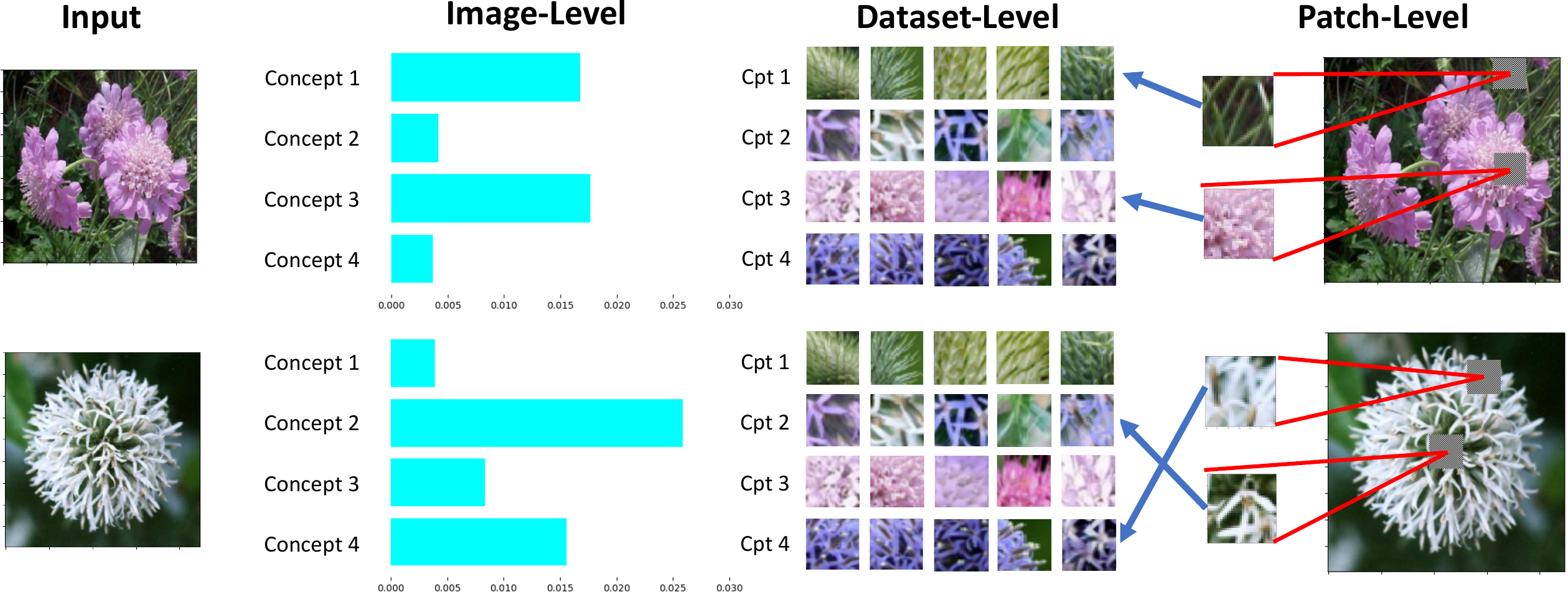} 
        \vskip -0.3cm
        \caption[width=0.8\linewidth]{PACE's three-level conceptual explanations on the \emph{Flower} dataset. \textbf{Dataset-Level}: PACE's top 4 dataset-level concepts (i.e., `Cpt 1' to `Cpt 4'); for each concept, we plot the top 5 patches with $\e_{mj}$ closest to $\muu_k$. 
        \textbf{Image-Level}: 
        Given an input image $m$, e.g., the top-left image, PACE's generated image-level explanation $\tha_m$ indicates a strong association with Concepts $1$ (\green{green} stem/leaves) and $3$ (\purple{purple} petal). This is consistent with the image's petal as the foreground and stem/leaves as the background. 
        \textbf{Patch-Level}:       
        For an input image $m$, e.g., the top-left image, PACE's $\ph_{mj}$ identifies the top concepts for patch $j$. The patch at the top (\green{green} stem/leaves) is associated with Concept $1$ (containing similar appearance patches across the dataset); the middle patch (\purple{purple} petal) is linked to Concept $3$ (containing patches of other pincushion flower petal across the dataset).}
        \label{fig:flower_qualitative}
       \vskip -0.4cm
\end{figure*}

\subsection{Quantitative Results}\label{sec:quantitative}
\tabref{tab:real_world_compare} shows the quantitative results for our PACE and the baselines for the desiderata in \defref{def:concept} across one synthetic dataset (\emph{Color}) and three real-world datasets (\emph{Flower}, \emph{Cars}, and \emph{CUB}). 
{For a detailed discussion on Multi-level Structure and Parsimony, please refer to \textbf{\appref{app:implement}}.}
Below we discuss Faithfulness, Stability, and Sparsity in detail.  
\textbf{\emph{Color}.} 
On the \emph{Color} dataset, PACE distinctly surpasses other leading models, as detailed in~\tabref{tab:real_world_compare}. PACE achieves perfect faithfulness ($1.00$) and the best stability score ($0.20$), demonstrating consistency in its explanations. It leads in sparsity ($0.97$), delivering focused and clear explanations. 

\textbf{\emph{Flower}, \emph{Cars}, and \emph{CUB}.} 
Our evaluation on three real-world datasets -- \emph{Flower}, \emph{Cars}, and \emph{CUB} -- reveals PACE's significant advantages over established baselines across various desiderata. As shown in~\tabref{tab:real_world_compare}, PACE consistently registers the highest faithfulness scores ($0.80$ on \emph{Flower}, $0.50$ on \emph{Cars}, and $0.56$ on \emph{CUB}), reflecting its superior precision in mirroring the model's decision-making process; note that both \emph{Cars} and \emph{CUB} contain a large number of classes ($196$ and $200$); therefore, a linear faithfulness score of $0.50$ is already very high. As mentioned in~\secref{sec:metrics}, one can always fit more complex models (e.g., nonlinear models such as a two-layer neural networks) to evaluate nonlinear faithfulness; for simplicity, we focus on linear faithfulness in this paper. 
Its stability scores (lower is better) on these three datasets are $0.12$, $0.05$, and $0.05$, respectively, illustrating its resilience to input perturbations. In terms of sparsity, PACE is highly competitive, achieving second-best results and thus providing succinct, pertinent explanations. 

\begin{table}[t]
\vskip -0.2cm
\caption{Average results across all four datasets in terms of faithfulness, stability, and sparsity for different methods. The best results are marked with \textbf{bold face}. {$\uparrow/\downarrow$ indicates higher/lower is better, respectively.}}
\vskip  0.1 cm
\label{tab:avg_compare}
\centering
  \resizebox{0.45\textwidth}{!}{%
\begin{tabular}{l|ccccccc}
\toprule
{Desiderata}& {SHAP} & {LIME} & {Saliency} & {AGI} & CRAFT & {PACE} \\

\midrule
Faithfulness $\boldsymbol{\uparrow}$ & 0.44 & 0.16  & 0.64  & 0.59 & 0.15   & \textbf{0.72} \\
Stability $\boldsymbol{\downarrow}$ & 2.02  & 1.57 & 0.43  & 1.74  & 1.85  & \textbf{0.11} \\
Sparsity $\boldsymbol{\uparrow}$ & 0.23 & 0.55  & 0.00 & 0.04  & 0.23  & \textbf{0.64} \\
\bottomrule
\end{tabular}
}
\vskip -0.5cm
\end{table}

\textbf{Average Performance Across Datasets.} 
\tabref{tab:avg_compare} shows the average performance across all four datasets in terms of faithfulness, stability, and sparsity. 
PACE consistently leads in faithfulness ($0.72$), stability ($0.11$), and sparsity ($0.64$). Compared to other models, its improvements are substantial, enhancing faithfulness by $0.08\sim 0.57$, stability by $0.32\sim 1.91$, and sparsity by $0.09\sim 0.64$, verifying PACE's effectiveness in terms of trustworthy explanations. 


\subsection{Qualitative Analysis}\label{sec:qualitative}


\textbf{\emph{Color}.} 
\figref{fig:color_qualitative} illustrates PACE's three-level explanations on the \emph{Color} dataset, where the ViT correctly predicts the top and bottom input images as \emph{Class 1} (\green{green}/\blue{blue} colors) and \emph{Class 0} (images with \red{red}/\yellow{yellow} colors), respectively. 
\begin{itemize}
\item \textbf{Dataset-Level Explanation:} The dataset-level column shows PACE's top 4 dataset-level concepts (for each concept, we plot the top 3 patches with $\e_{mj}$ closest to $\muu_k$); they are consistent with the 4 primary colors in the dataset (see~\secref{sec:datasets}), verifying the PACE's effectiveness. 
\item \textbf{Image-Level Explanation:} Given an input image $m$, PACE infers the image-level concepts. For example, given the input image in the top row of \figref{fig:color_qualitative}, PACE's generated image-level explanation $\theta_m$ indicates a strong association with Concept $3$ (\textcolor{green}{green}) and Concept $4$ (\textcolor{blue}{blue}). This is consistent with the color distribution in the image $m$, predominantly \textcolor{blue}{blue} and less \textcolor{green}{green}.
\item \textbf{Patch-Level Explanation:} PACE also generated patch-level explanation. For the same image above, PACE's $\ph_{mj}$ identifies the top concepts for the selected patches; 
the \blue{blue} patch is associated with Concept $4$ (containing similar \textcolor{blue}{blue} patches across the dataset), while the \textcolor{green}{green} patch is linked to Concept $3$ (containing \textcolor{green}{green} patches across the dataset). 
\end{itemize}

\textbf{\emph{Flower}.} 
\figref{fig:flower_qualitative} demonstrates PACE's three-level explanations on the \emph{Flower} dataset. 
\begin{itemize}
\item \textbf{Dataset-Level Explanation:} The dataset-level column shows the top 4 dataset-level concepts from our PACE, each with unique shapes, texture, and colors. 
\item \textbf{Image-Level Explanation:} Given an input image $m$, PACE infers the image-level concepts. For example, given the input image in the top row of \figref{fig:flower_qualitative}, PACE's generated image-level explanation $\tha_m$ indicates a strong association with Concepts $1$ (\green{green} stem/leaves) and $3$ (\purple{purple} petal). This is consistent with the image's petal as the foreground and stem/leaves as the background. 
\item \textbf{Patch-Level Explanation:} PACE also generated patch-level explanation. For the same image $m$ above, PACE's $\ph_{mj}$ identifies the top concepts for patch $j$. The patch at the top (\green{green} stem/leaves) is associated with Concept $1$, comprising similar appearance patches across the dataset; the middle patch (\purple{purple} petal) is linked to Concept $3$, which includes patches of other pincushion flower petal across the dataset.
\end{itemize}


See~\appref{app:qualitative} for further qualitative analysis on more real-world datasets. 

\section{Conclusion}
In this paper, we identify five desiderata \emph{faithfulness}, \emph{stability}, \emph{sparsity}, \emph{multi-level structure}, and \emph{parsimony} when generating trustworthy concept-level explanations for ViTs. {We develop the first general method, PACE, that is compatible with any transformer variants and satisfies these desiderata.} Through both quantitative and qualitative evaluations, our method demonstrates superior performance in explaining post-hoc ViT predictions via visual concepts, outperforming state-of-the-art methods across various  datasets. As a limitation, our approach assumes a fixed number of concepts (Similar to existing methods). Therefore future work could focus on developing PACE into a non-parametric explainer that automatically determines the number of concepts. Another limitation is that our approach requires access to hidden states and attention weights from the layers inside ViTs; we argue that this is an advantage because it allows our PACE to interpret vision foundation models' internals thoroughly rather than simply their output superficially. 



\section*{Acknowledgements}
We extend our heartfelt thanks to Akshay Nambi and Tanuja Ganu from Microsoft Research for their invaluable suggestions, which greatly improved the presentation of this paper. We are grateful for the support from Microsoft Research AI \& Society Fellowship, NSF Grant IIS-2127918, and Amazon Faculty Research Award. Additionally, we thank the reviewers and the area chair/senior area chair for their insightful feedback and for recognizing the novelty and contributions of our work. 
We thank the Center for AI Safety (CAIS) for making computing resources available for this research. 
\section*{Impact Statement}    
{This paper presents work whose goal is to advance the field of Machine Learning. There are many potential societal consequences of our work, none which we feel must be specifically highlighted here.}
\bibliography{main}
\bibliographystyle{icml2024}

\newpage
\appendix
\onecolumn

\section{Implementation Details}\label{app:implement}
In this section, we provide implementation details of PACE. 

\textbf{\emph{Color} Dataset Generation.} We constructed \emph{Color} as a synthetic dataset with clear definition of 4 concepts (\red{red}/\yellow{yellow}/\green{green}/\blue{blue}). It contains two image classes: \emph{Class 0} (images with \red{red}/\yellow{yellow} colors) and \emph{Class 1} (\green{green}/\blue{blue} colors), both against a \emph{black} background (see example images in~\figref{fig:color_samples}). Images are initially created at a $2\times2$ resolution, where each pixel samples color from (\red{red}/\yellow{yellow},\green{green}/\blue{blue},black), and are subsequently up-sampled with gradual color shift to $224\times224$ for ViT inputs. We introduce random Gaussian noise to each image. The dataset includes 1000 images per class, with a split of 800 training and 200 test samples.

\begin{table}[h]
 \vskip -0.5cm
      \caption{Dataset statistics, i.e., the number of train/test images ($M_{train}/M_{test}$), the number of classes $N$, and the number of patches $J$ per image.}\label{tab:dataset}
        \footnotesize
      \centering
      \begin{tabular}{lcccc}
        \toprule
        \textbf{Dataset}  & $M_{train}$ & $M_{test}$  & $N$ & $J$   \\
        
       \midrule
        Color  & 1,600 &  400  &  2 & 197\\
        Flower &7,169 & 1,020 & 102 & 197 \\
        Cars & 8,144 & 8,041 & 196&  197  \\
        CUB & 5,994&5,794  & 200& 197\\
        \bottomrule
      \end{tabular}
\end{table}
{\tabref{tab:dataset}} provides statistics for the COLOR dataset along with three additional real-world datasets.

\textbf{Experimental Setup.}
Following the approach outlined in~\cite{chen2020simclr}, we implement perturbation {described in~\defref{def:concept}} based on their augmentation algorithms, as demonstrated by the following code snippet:
{\small\begin{verbatim}
   contrast_transforms = transforms.Compose([transforms.RandomHorizontalFlip(),
                                          transforms.RandomResizedCrop(size=size),
                                          transforms.RandomApply([
                                              transforms.ColorJitter(brightness=0.5,
                                                                     contrast=0.5,
                                                                     saturation=0.5,
                                                                     hue=0.1)
                                          ], p=0.8),
                                          transforms.RandomGrayscale(p=0.2),
                                          transforms.GaussianBlur(kernel_size=9),
                                          transforms.Normalize((0.5,), (0.5,))
                                         ])  
\end{verbatim}}
We also utilize in-batch negative examples according to~\cite{chen2020simclr}. 
We implemented and trained using PyTorch~\cite{paszke19pytorch} on an A5000 GPU with 24GB of memory. The training duration does not exceed one day for all four datasets. We employ the Adam optimizer~\cite{kingma14adam} with initial learning rates ranging from $10^{-5}$ to $10^{-3}$, depending on the dataset.

From preliminary results, we observed that a smaller value for $K$ is inadequate for effectively learning significant image concepts. Conversely, a larger value for $K$ tends to lead to redundancy among the population of all concepts. {Consequently, we adhered to the baseline methodologies (e.g.,~\cite{fel2023craft}) by setting $K$ to 100 across all datasets. This number was chosen as it strikes an effective balance between capturing adequate detail and avoiding model overfitting.}

\textbf{{Baselines Methods.}}
{For the implementation of the baseline methods, we either utilize the original packages provided by the authors~\cite{lundberg2017shap,ribeiro2016lime,fel2023craft}, or implement their methods by referencing the authors' code~\cite{simonyan2013saliency,pan2021agi}.} To account for the stochastic nature of methods like those in~\cite{ribeiro2016lime, lundberg2017shap}, we perform multiple executions of these baseline methods, averaging scores across all runs. The frequency of repetition is contingent upon the time required to generate explanations. To balance efficiency and effectiveness, SHAP is executed 100 times, and LIME 10 times. In contrast, our PACE, being fully deterministic post-training, requires only a single inference on the test set.

{\textbf{Details on the Quantitative Analysis.}}
In the results shown in \tabref{tab:real_world_compare} in~\secref{sec:quantitative}, PACE offers \emph{Full} three-level conceptual explanations, encompassing dataset, image, and patch levels. In contrast, CRAFT \cite{fel2023craft} is limited to providing explanations at only the patch and image levels, lacking dataset-level insights and thereby exhibiting a \emph{Partial} multi-level structure. Other baseline models are unable to achieve this multi-level conceptual explanation. 
Both the baseline CRAFT and our PACE are inherently compatible with arbitrary number of concepts $K$, therefore enjoying better parsimony by setting $K=100$. This choice of $K$ is driven by the goal of maintaining a moderate dimension size while ensuring that concept activation possesses meaningful semantics. In contrast, other baselines' number of concepts $K$ is constrained to the hidden dimension of ViT embeddings, i.e., $K = 768$; they are therefore lacking in parsimony.


\textbf{Details on the Qualitative Analysis.}
For qualitative analysis, we visualize $2\times 2$ aggregated patches, chosen for their visibility and robustness against random noise. The aggregation only affects patch-level concepts, computed similarly to $\bar \z_m$ in~\eqnref{z_avg_def}. The mean of $\ph_{mj}$ approximates the patch-level concept for each aggregated patch $\hat \ph_{mj}$, computed as follows:
\begin{align}
\hat \ph_{m,u\cdot \frac{S}{2}+v} = \frac{1}{4}(\ph_{m, 2u\cdot S+2v}+\ph_{m, 2u\cdot S+2v+1}+\ph_{m, (2u+1)\cdot S+2v}+\ph_{m, (2u+1)\cdot S+2v+1}),
\end{align}
where $S$ is the number of rows and columns. 
Note that this aggregation is for visualization purposes only during \emph{qualitative} analysis and does not affect the quantitative results, dataset-level and image-level concepts, or the learning process.

\section{Sparsity versus Parsimony} \label{app:sparse_vs_pars}
As we discuss in~\defref{def:concept}, \emph{sparsity} is defined as the proportion of $\tha_m$'s entries nearing zero, i.e., $\frac{1}{K}\sum\nolimits_{k=1}^K\mymathbb{1}(\vert \tha_{mk}\vert<\epsilon)$, 
    with a small threshold $\epsilon>0$; \emph{parsimony} is defined as the minimal number of concepts $K$ to produce clear and simple explanations.  
While sparsity is an \emph{image-level} property, parsimony is \emph{dataset-level}.

\textbf{Example 1.} 
For example, first consider a dataset with four concepts and three images, $\I_1$, $\I_2$, and $\I_3$:

\begin{itemize}
\item On the image level,
$\I_1$: $\tha_1=(1,0,0,0)$.
$\I_2$: $\tha_2=(0,1,0,0)$.
$\I_3$: $\tha_1=(0,0,1,0)$.
\item On the dataset level, $\muu_m=\e_m (1\leq m \leq 3)$, $\muu_4=\frac{1}{3}(\muu_1+\muu_2+\muu_3)$. 
\end{itemize}

According to~\defref{def:concept}, these image-level concepts satisfies \emph{sparsity}; however, the dataset-level concept does not satisfy \emph{parsimony}, since the last concept center $\muu_4$ is redundant.

\textbf{Example 2.} 
Next, we consider a dataset with three concepts and three images, $\I_1$, $\I_2$, and $\I_3$:

\begin{itemize}
\item On the image level,
$\I_1$: $\tha_1=(0.5,0.5,0)$.
$\I_2$: $\tha_2=(0.5,0,0.5)$.
$\I_3$: $\tha_1=(0,0.5,0.5)$.
\item On the dataset level, $\muu_1=\frac{1}{2}(\e_1+\e_2)$, $\muu_2=\frac{1}{2}(\e_1+\e_3)$, and $\muu_1=\frac{1}{2}(\e_2+\e_3)$. 
\end{itemize}

According to~\defref{def:concept}, These image-level concepts apparently does not satisfy \emph{sparsity}; however, the dataset-level concept satisfies \emph{parsimony}, because there are no redundant concepts. 
Therefore, in our paper, \emph{sparsity and parsimony, though related, are distinct and non-interchangeable properties}.



 \begin{figure}[!t]
        \centering
        \includegraphics[width = 0.95\textwidth]{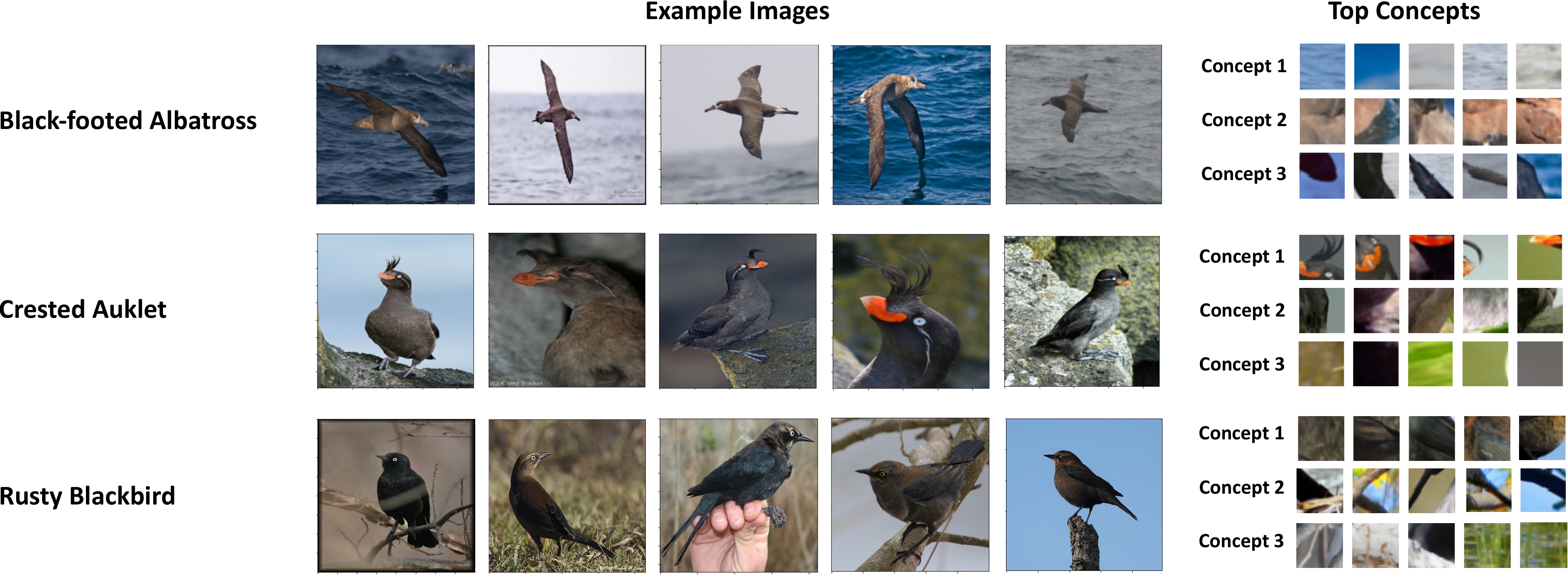} 
        \caption[width = 0.8\linewidth]{PACE's dataset-level conceptual explanations for classes \textbf{Black-footed Albatross}, \textbf{Crested Auklet}, and \textbf{Rusty Blackbird} in the \emph{CUB} dataset. 
        For each class, we show PACE's top 3 dataset-level concepts; for each Concept $k$, we show the top 5 patches with their associated embeddings $\e_{mj}$ closest to the concept center $\muu_k$.}
        \label{fig:cub_qualitative}
\end{figure}

\section{More Qualitative Results} \label{app:qualitative}
\begin{figure}[!t]
        \centering
        \includegraphics[width = 0.95\textwidth]{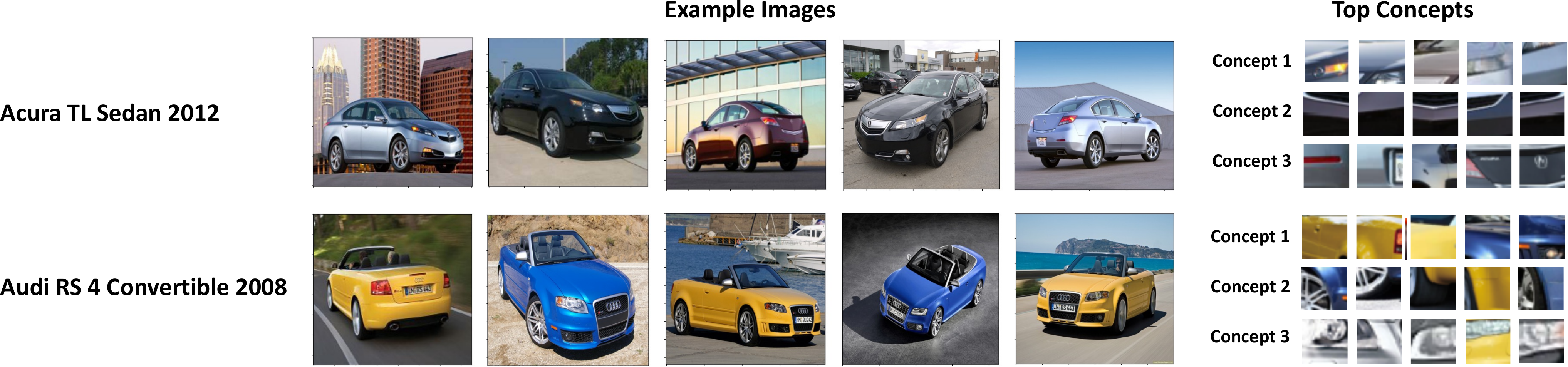} 
        \caption[width = 0.8\linewidth]{PACE's dataset-level conceptual explanations for classes $\textbf{Acura TL Sedan 2012}$ and $\textbf{Audi RS 4 Convertible 2008}$ in the \emph{Cars} dataset. 
        For each class, we show PACE's top 3 dataset-level concepts; for each Concept $k$, we show the top 5 patches with their associated embeddings $\e_{mj}$ closest to the concept center $\muu_k$.}
        \label{fig:cars_qualitative}
\end{figure}

In \figref{fig:cub_qualitative} and \figref{fig:cars_qualitative}, we present the top three concepts for several distinct classes in the \emph{CUB} and \emph{Cars} datasets, respectively. Each concept is illustrated with the top five patches, providing dataset-level explanations.

\textbf{Results on \emph{CUB}.} 
\figref{fig:cub_qualitative} shows PACE's dataset-level conceptual explanations for the \emph{CUB} dataset's classes $\textbf{Black-footed Albatross}$, $\textbf{Crested Auklet}$, and $\textbf{Rusty Blackbird}$. For instance, the class $\textbf{Black-footed Albatross}$ encompasses three predominant concepts: Concept 1 (\emph{Ocean Background}), Concept 2 (\emph{Brown Feather}), and Concept 3 (\emph{Long Wing}). The accompanying top five patches exemplify PACE's conceptual explanations, highlighting critical dataset-level concepts such as the habitat (\emph{Ocean}), distinctive texture (\emph{Brown Feather}), and characteristic posture (\emph{Long Wing}) crucial for classifying $\textbf{Black-footed Albatross}$. Similarly, the class  $\textbf{Crested Auklet}$ is distinguished by concepts such as Concept 1 (\emph{Orange Beak}), Concept 2 (\emph{Grey Feather}), and Concept 3 (\emph{Rocks/Moss}); similarly, the class  $\textbf{Rusty Blackbird}$ is distinguished by Concept 1 (\emph{Rusty Feather}), Concept 2 (\emph{Tail}), and Concept 3 (\emph{Grass/Branch}). These findings reveal that distinct bird classes are each linked to unique body characteristics, such as color, shape, and texture, as well as specific habitats.

\textbf{Results on \emph{Cars}.} 
\figref{fig:cars_qualitative} shows PACE's dataset-level conceptual explanations for the \emph{Cars} dataset's classes, such as \textbf{Acura TL Sedan 2012} and \textbf{Audi RS 4 Convertible 2008}. For example, the \textbf{Audi RS 4 Convertible 2008} class features three prominent concepts: Concept 1 (\emph{Front Light}), Concept 2 (\emph{Grill}), and Concept 3 (\emph{Rear}). The top five patches representing these concepts indicate that design elements like the front light (\emph{Front Light}), grill pattern (\emph{Grill}), and rear features (\emph{Rear}) are essential for classifying an image as class \textbf{Acura TL Sedan 2012}. Moreover, the \textbf{Audi RS 4 Convertible 2008} class is defined by concepts such as Concept 1 (\emph{Streamline}), Concept 2 (\emph{Tire/Fender}), and Concept 3 (\emph{Front Light}), suggesting that car classes differ in design aspects such as shape and color.

\textbf{Remark.} 
In summary, these results showcase PACE's proficiency in identifying crucial dataset-level concepts across different classes, utilizing patch representations within the ViT framework. Notably, this process, once the model is trained, involves deducing top concepts for each class via inference, eliminating the need for retraining or finetuning. This approach is both efficient and effective compared to methods like CRAFT~\cite{fel2023craft} that require training for each individual class.




\section{Details on Inferring Concepts} \label{app:explain_g}
In this section, we discuss $g(\cdot)$ defined in~\secref{sec:notations} in detail.

In PACE, $g(\cdot)$ is implemented as an inference process on our Probabilistic Graphical Model (PGM), as shown in~\figref{fig:pgm}. One can see $g(\cdot)$ as a function that
\begin{itemize}
    \item[(1)] takes the observed variables $\e_m, \e'_m, \hat \y_m, \a_m, \a'_m, y_s$ as inputs,
    \item[(2)] goes through the {learning stage} and the {inference stage} discussed in~\secref{sec:learning} and~\secref{sec:inference}, and
    \item[(3)] outputs $\tha_m$, the image-level concept explanations for each image $m$. 
\end{itemize}

As shown in~\figref{fig:overview}, $g(\cdot)$ refers to the PACE model (the gray box on the top right). It takes as inputs the patch embeddings $\e_m$, the attention weights $\a_m$ (which can be computed given the ViT's parameters $P$), and the ViT's predicted label $\hat \y_m$; it then outputs the image-level explanations $\tha$, i.e., $\tha_m = g(\e_m, \hat \y_m, \a_m)$. 

Besides the image-level concept explanations $\tha_m$, PACE also produces the dataset-level explanations $\muu_k$ and $\Si_k$ (where $k=1, \dots, K$) as well as patch-level explanations $\ph_{mj}$ for patch $j$ of image $m$).

$g(\cdot)$ is represented by the entire~\figref{fig:pgm}, except for the dashed box (with the text ``ViT" inside). For example, during the {inference stage}, PACE will 
\begin{itemize}
    \item[(1)] be given the \emph{global} parameters $\muu_k$ and $\Si_k$ (where $k=1, \dots, K$) obtained from the {learning stage}, 
    \item[(2)] treats the patch embeddings $\e_{m}$, the attention weights $\a_{m}$ (which can be computed given the ViT's parameters $P$), and the ViT's predicted label $\hat \y_m$ as observed variables,
    \item[(3)] and then, for a new image $m$, infer the \emph{local} parameters, i.e., 
    \begin{itemize}
        \item[(a)] the image-level concepts (explanations) $\tha_m$, which is parameterized by $q(\tha_m|\gamm_m)$ and 
        \item[(b)] patch-level concepts (explanations) $\z_{mj}$, which is parameterized by $q(\z_{mj}|\ph_m)$.
    \end{itemize}
\end{itemize}

These are called \emph{local} parameters because each image has its own $\tha_m$ and $\ph_m$.

\section{Theoretical Analysis}
\label{app:proof}
We provide the following proof of~\thmref{thm:MI_bound}. For convenience, let $\Omega = (\muu_{k=1}^K, \Sigma_{k=1}^K)$. We then introduce a helper joint distribution of the variables $\e_m$ and $\tha_m, \ph_m$, $s(\e_m, \tha_m, \ph_m)=p(\e_m)q(\tha_m, \ph_m|\e_m)$.

According to the definition of ELBO of~\secref{sec:inference}, in~\eqnref{eq:full_elbo} and~\eqnref{eq:full_elbo_expand}, we only need to prove that
\begin{align}\label{eq:MI_bound_app}
 LHS =~& L_e + L_f + L_s\nonumber\\
\leq~& {I}(\e_m; \tha_m,\ph_m) + I(\hat \y_m;\ph_m) 
     + I(\ph_m;\ph'_m) 
     +  C.
\end{align}

We split the proof into the following three separate part:

\textbf{(1) The bound of $L_e$.} We have that
\begin{align}
L_e = \EB_{p(\e_m)}[\EB_{q}[\log p(\e_m|\Omega, \tha_m, \ph_m)]] + \EB_{q}[\log q(\tha_m, \ph_m|\Omega)].
\end{align}
Since $\EB_{q}[\log q(\tha_m, \ph_m|\Omega)] \le 0$, we are going to prove that 
\begin{align}
L_e\leq\EB_{p(\e_m)}[\EB_{q}[\log p(\e_m|\Omega, \tha_m, \ph_m)]] \le I_s(\e_m;\tha_m, \ph_m) - H(\e_m).
\end{align}
In fact,
\begingroup\makeatletter\def\f@size{7.5}\check@mathfonts
\begin{align}
\EB_{p(\e_m)}[\EB_q[\log p(\e_m|\tha_m, \ph_m,\Omega)]] &\le 
\EB_{p(\e_m)}[\EB_q[\log p(\e_m|\tha_m, \ph_m)]] \nonumber\\ 
&=\EB_{p(\e_m)}[\EB_q[\log\frac{q(\e_m|\tha_m, \ph_m)}{p(\e_m)}\frac{p(\e_m)p(\e_m|\tha_m, \ph_m)}{q(\e_m|\tha_m, \ph_m)}]]\nonumber\\
&=\EB_{p(\e_m)}[\EB_q[\log\frac{q(\e_m|\tha_m, \ph_m)}{p(\e_m)}]]
+\EB_{p(\e_m)}[\EB_q[\log p(\e_m)]] + \EB_{p(\e_m)}[\EB_{q}[\log\frac{p(\e_m|\tha_m, \ph_m)}{q(\e_m|\tha_m, \ph_m)}]]\nonumber\\
&=I_s(\e_m;\tha_m, \ph_m) - H(\e_m) - \EB_{q}[KL(q(\e_m|\tha_m, \ph_m)|p(\e_m|\tha_m, \ph_m))]\nonumber\\
&\le I_s(\e_m;\tha_m, \ph_m) - H(\e_m) - 0,
\end{align}
\endgroup
where $H(\e_m)$ is a constant.

\textbf{(2) The bound of $L_f$.} With the constraint $-\1 \leq \et_n\leq\1 (1\leq n\leq N)$, we have that
\begin{align}
    L_f =~& \mathbb{E}_{q}[\log p(\hat\y_m|\bar\z_m,\mH)] \nonumber\\
     \approx ~& \sum\nolimits_{n=1}^{N} \hat y_{mn} (\et^T_{n}  \bar \ph_m) 
     -   \log (\sum\nolimits_{n=1}^{N} \exp (\et^T_{n} \bar\ph_m))\nonumber\\
     \leq ~& \sum\nolimits_{n=1}^{N} \hat y_{mn} (\et^T_{n}  \bar \ph_m) \nonumber\\
     \leq ~& \sum\nolimits_{n=1}^{N} \sum\nolimits_{k=1}^{K}  \hat y_{mn}\bar \phi_{mk} 
     \nonumber \\
     \leq~&\sum_{\hat\y_m} \sum_{\ph_m} p(\hat \y_{m},\bar \ph_{m} )\log\frac{p(\hat \y_{m},\bar \ph_{m} )}{p(\hat \y_{m})p(\bar \ph_{m}) } + C_1\nonumber\\
     =~& I(\hat \y_m;\ph_m) + C_1,
\end{align}
where $C_1$ is a constant. 

\textbf{(3) The bound of $L_s$.} With the constraint $\0\leq\bet\leq\1$, we have that
\begin{align}
    L_s =~& \EB_{q}[\log p(y_s=1\vert \bar \z_{1:M},\bar \z'_m, \bet)] 
    \nonumber\\
     \approx~& \bet^T (\bar\ph_{m}\circ\bar\ph'_{m})
     - \log (\sum\nolimits_{f\in \FM} \exp(\bet^T (\bar\ph_{m}\circ\bar\ph_{f})) ) \nonumber\\
     \leq ~& \bet^T (\bar\ph_{m}\circ\bar\ph'_{m})\nonumber\\
     \leq ~& \bar\ph_{m}\cdot\bar\ph'_{m} \nonumber \\
     \leq~&\sum_{\ph_m} \sum_{\ph'_m} p(\ph_m, \ph'_{m} )\log\frac{p(\ph_m, \ph'_{m} )}{p(\ph_m)p(\ph'_{m}) } + C_2\nonumber\\
     =~& I(\ph_m;\ph'_m) +C_2,
\end{align}
where $C_2$ is a constant. 

Combining $(1\sim3)$ above concludes the proof.

\section{Details on Learning PACE} 

\subsection{Derivations of ELBO}\label{app:derivation}
\label{sec:app_update_rules}
\textbf{Inferring $\z_m$.}
According to~\eqnref{z_avg_def},
\begin{align}
    \bar \z_m = \frac{1}{J}\sum\nolimits_{j=1}^J \z_{mj}, 
\end{align}
where $\z_{mj}$ can be approximate by a variational distribution parameterized by $\ph_{mj}$:
\begin{align}
q(\z_{mj} \mid \ph_{mj}) = \text{Categorical}(\z_{mj} \mid \ph_{mj}),
 \end{align}
which indicates that
\begin{align}
    \EB[\z_{mj}] = \ph_{mj}.
\end{align}
Therefore, we have 
\begin{align}
    \EB[\bar\z_m] = \frac{1}{J}\sum\nolimits_{j=1}^J \EB[\bar\z_{mj}] 
    = \frac{1}{J}\sum\nolimits_{j=1}^J \ph_{mj} = \bar \ph_m.
\end{align}
Hence, we have
\begin{align}
    \bar \z_m \approx \bar \ph_{m}.
\end{align}

\textbf{Computing $L_e$.}
We can expand the ELBO in~\eqnref{eq:elbo} as: 

    \begin{align} \label{eq:elbo_expand}
    L_e =~&  \log\Gam(\sum_{k=1}^K\alpha_k)-\sum_{k=1}^K\log\Gam(\alpha_k) + 
    \sum_{k=1}^K(\alpha_k-1)(\Psi(\gamm_{mk})-\Psi(\sum_{k'=1}^K\gamm_{k'}))\nonumber\\
    &+\sum_{k=1}^K\phi_{mjk}(\Psi(\gamm_{mk})-\Psi(\sum_{k'=1}^K\gamm_{mk'}))\nonumber\\
    &+\sum_{k=1}^K \phi_{mjk} a_{mj} \{-\frac{1}{2}(\e_{mj}-\muu_k)^T\Si_k^{-1}(\e_{mj}-\muu_k)
    -\log[(2\pi)^{d/2} \vert \Si_k\vert^{1/2}]\} \nonumber\\
    &-\log \Gam(\sum_{k=1}^K \gamm_{mk}) + \sum_{k=1}^K \log\Gam(\gamm_{mk}) - \sum_{k=1}^K (\gamm_{mk}-1)(\Psi(\gamm_{mk})-\Psi(\sum_{k'=1}^K\gamm_{mk'}))\nonumber\\
    &-\sum_{k=1}^K \phi_{mjk}\log \phi_{mjk}.
    \end{align}

We can interpret the meaning of each term of $L_e$ as follows:
\begin{itemize}
    \item The sum of the first and the fourth terms, namely ${\EB}_q[\log p(\tha_m|\alp)] - {\EB}_q[\log q(\tha_m)]$, is equal to $-KL(q(\tha_m)|p(\tha_m|\alp))$, which is the negation of KL Divergence between the variational posterior probability $q(\tha_m)$ and the prior probability $p(\tha_m|\alp)$ of the topic proportion $\tha_m$ for document $m$. Therefore maximizing the sum of these two terms is equivalent to minimizing the KL Divergence $KL(q(\tha_m)|p(\tha_m|\alp))$; this serves as a regularization term to make sure the inferred $q(\tha_m)$ is close to its prior distribution $p(\tha_m|\alp)$.  

 \item Similarly, the sum of the second and the last terms (ignoring the summation over the word index $j$ for simplicity), namely ${\EB_q}[\log p(z_{mj}|\tha_m)] - {\EB_q}[\log q(z_{mj})]$ is equal to $-KL(q(z_{mj})|p(z_{mj}|\tha_m))$, which is the negation of the KL Divergence between the variational posterior probability $q(z_{mj})$ and the prior probability $p(z_{mj}|\tha_m)$ of the word-level topic assignment $z_{mj}$ for word $j$ of document $m$. Therefore maximizing the sum of these two terms is equivalent to minimizing the KL Divergence $KL(q(z_{mj})|p(z_{mj}|\tha_m))$; this serves as a regularization term to make sure the inferred $q(z_{mj})$ is close to its ``prior" distribution $p(z_{mj}|\tha_m)$. 

\item The third term $\EB_q [\log p(\e_{mj}|z_{mj},\muu_{z_{mj}}, \Si_{z_{mj}})]$ is to maximize the log likelihood $p(\e_{mj}|z_{mj},\muu_{z_{mj}}, \Si_{z_{mj}})$ of every contextual embedding $\e_{mj}$ (for word $j$ of document $m$) conditioned on the inferred $z_{mj}$ and the parameters $(\muu_{z_{mj}}, \Si_{z_{mj}})$. 
\end{itemize}

\textbf{Computing $L_f$.}
\eqnref{eq:faithful} is derived from employing Taylor's expansion to~\eqnref{eq:prob_class}:
\begin{align}
    L_f =~& \mathbb{E}_{q}[\log p(\hat\y_m|\bar\z_m,\mH)] \nonumber \\
     =~& \sum\nolimits_{n=1}^{N} \hat y_{mn} (\et^T_{n}  \bar \ph_m) 
     -  \EB_{q} [\log (\sum\nolimits_{n=1}^{N} \exp (\et^T_{n} \bar\z_m))] \nonumber\\
     \approx ~& \sum\nolimits_{n=1}^{N} \hat y_{mn} (\et^T_{n}  \bar \ph_m) 
     -   \log (\sum\nolimits_{n=1}^{N} \exp (\et^T_{n} \bar\ph_m + (1/2)\et_n^T \S_m\et_n)),     
\end{align}
where $\S_m$ is the covariance matrix of $\bar\z_m$.

We will see that for any entry of $\S_m$, i.e. $\forall x,y \in \{1,2,...,K\}$, we have 
\begin{align} \label{eq:cov_bound}
0 \leq \S_{m,xy} \leq \frac{1}{J^2}.
\end{align}
In our setting, the number of patches $J$ in each image satisfies $J>100$, hence $\S_{m,xy}$ is very close to zero.

We compute $\S_m$ by definition:
\begin{align}\label{eq:cov_z_mul}
    \S_{m,xy}  = ~& Cov[\bar \z_{m} \bar \z_{m'}]_{x,y} \nonumber \\
    =~& \EB[(\bar z_{mx} \bar z_{m'x}-\EB[\bar z_{mx} \bar z_{m'x}])(\bar z_{my} \bar z_{m'y}-\EB[\bar z_{my} \bar z_{m'y}])]  \nonumber\\
    =~& \EB[(\bar z_{mx} \bar z_{m'x}-\EB[\bar z_{mx}]\EB[ \bar z_{m'x}]])(\bar z_{my} \bar z_{m'y}-\EB[\bar z_{my}]\EB[ \bar z_{m'y}]])]  \nonumber\\
    =~& \EB[(\bar z_{mx} \bar z_{m'x}-\bar \phi_{mx}\bar \phi_{m'x})(\bar z_{my} \bar z_{m'y}-\bar \phi_{my}\bar \phi_{m'y})]  \nonumber \\
    = ~& \EB[\bar z_{mx} \bar z_{m'x}\bar z_{my} \bar z_{m'y}] - \bar\phi_{my}\bar \phi_{m'y}\EB[\bar z_{mx} \bar z_{m'x}] - \bar\phi_{mx}\bar \phi_{m'x}\EB[\bar z_{my} \bar z_{m'y}] + \bar\phi_{mx}\bar \phi_{m'x}\bar\phi_{my}\bar \phi_{m'y} \nonumber\\
    = ~& \EB[\bar z_{mx} \bar z_{my}] \EB[\bar z_{m'x} \bar z_{m'y}] - \bar\phi_{mx}\bar \phi_{m'x}\bar\phi_{my}\bar \phi_{m'y}. 
\end{align}
We then consider two different cases:

\textbf{Case (1): $x=y$.}

Then we have that
\begin{align}
    Cov[\bar \z_{m} \bar \z_{m'}]_{x,y}  = ~& \EB[\bar z_{mx}^2] \EB[\bar z_{m'y} ^2] - \bar\phi_{mx}^2\bar\phi_{m'y}^2 \nonumber\\
    =~& \bar\phi_{mx}^2\bar\phi_{m'y}^2 - \bar\phi_{mx}^2\bar\phi_{m'y}^2 \nonumber\\
    =~& 0.
\end{align}

\textbf{Case (2): $x\neq y$.}

Note that
\begin{align}
    \bar z_{mx} \bar z_{my} = \frac{1}{J} \sum_j z_{mjx}\cdot  \frac{1}{J} \sum_j z_{mjy}.
\end{align}

Given that $\z_{mj}$ is a one-hot vector, we have
\begin{align}
    z_{mjx} \cdot z_{mjy} = 0.
\end{align}
Hence, we have
\begin{align}
    \EB[\bar z_{mx} \bar z_{my}] = ~& \EB[\bar z_{mx}]\EB[\bar z_{my}] - \frac{1}{J^2} (\sum_{j=1}^J \EB[z_{mjx}]\EB[z_{mjy}]) \nonumber\\
    = ~& \bar \phi_{mx}\bar \phi_{my} - \frac{1}{J^2} \sum_{j=1}^J\phi_{mjx}\phi_{mjy}.
\end{align}
Therefore, we have
\begin{align}
    Cov[\bar \z_{m} \bar \z_{m'}]_{x,y} =~& (\bar \phi_{mx}\bar \phi_{my} - \frac{1}{J^2} \sum_{j=1}^J \phi_{mjx}\phi_{mjy})
    (\bar \phi_{m'x}\bar \phi_{m'y} - \frac{1}{J^2} \sum_{j=1}^J \phi_{m'jx}\phi_{m'jy})
    - \bar\phi_{mx}\bar \phi_{m'x}\bar\phi_{my}\bar \phi_{m'y} \nonumber\\
    = ~&  \frac{1}{J^4} \sum_{j=1}^J \phi_{mjx}\phi_{mjy}\sum_{j=1}^J \phi_{m'jx}\phi_{m'jy}.
\end{align}
Since $\0\leq\ph_{mj}, \ph'_{mj}\leq \1$, we have that 
\begin{align}
    0\leq \S_{m,xy}  = Cov[\bar \z_{m} \bar \z_{m'}]_{x,y} \leq \frac{1}{J^2}.
\end{align}

In summary, we can see that in either case,~\eqnref{eq:cov_bound} holds. 

Therefore we have
\begin{align} \label{eq:faithful_simplified}
    L_f  \approx & \sum\nolimits_{n=1}^{N} \hat y_{mn} (\et^T_{n}  \bar \ph_m) 
     -   \log (\sum\nolimits_{n=1}^{N} \exp (\et^T_{n} \bar\ph_m)).
\end{align}

\textbf{Computing $L_s$.}
Similarly, by employ Taylor's expansion of~\eqnref{eq:prob_indicator}, as well as~\eqnref{eq:cov_bound}, we have that
\begin{align}\label{eq:stable_simplified}
    L_s =~& \EB_{q}[\log p(y_s=1\vert \bar \z_{1:M},\bar \z'_m, \bet)] 
 \nonumber\\
    =~&\bet^T (\bar\z_{m}\circ\bar\z'_{m})
     -  \EB_{q} [\log (\sum\nolimits_{f\in \FM} \exp(\bet^T (\bar\z_{m}\circ\bar\z_{f})) )] \nonumber\\
     \approx ~& \bet^T  \bar \ph_m \bar\ph'_m
     -   \log (\sum\nolimits_{f\in \FM}\exp (\bet^T(\bar\ph_m\bar\ph_f )+ (1/2)\bet^T\S_{m}\bet)  )     
     \nonumber\\
     \approx~& \bet^T (\bar\ph_{m}\circ\bar\ph'_{m})
     - \log (\sum\nolimits_{f\in \FM} \exp(\bet^T (\bar\ph_{m}\circ\bar\ph_{f})) ),
\end{align}
where the parameter $\bet$ is learned jointly by gradient-based optimization algorithms, such as Adam.

\subsection{Update Rules} \label{app:update}

\subsubsection{Inference}\label{app:inference}

\textbf{Derivative of $L_e$.}
Taking the derivative of the $L_e$ in~\eqnref{eq:elbo} with respect to $\phi_{mjk}$ and setting it to zero, we obtain the update rule for $\phi_{mjk}$:
\begin{align}
    \phi_{mjk} \propto ~& \frac{1}{\vert \Si_k\vert^{1/2}} 
      \exp[\Psi(\gamma_{mk})-\Psi(\sum\nolimits_{k'=1}^K \gamma_{k'}) \nonumber\\
     - &\frac{1}{2}a_{mj}(\e_{mj}-\muu_k)^T\Si_k^{-1}(\e_{mj}-\muu_k)].
\end{align}

\textbf{Derivative of $L_f$.}
The log-sum term of~\eqnref{eq:faithful_simplified} is intractable. To address this, with Taylor's Expansion, we have that
\begin{align}
\log (\sum\nolimits_{n=1}^{N} \exp (\et^T_{n} \bar\ph_m))  \approx~&
\log (\sum\nolimits_{n=1}^{N} \exp (\et^T_{n} \bar\ph^{(0)}_{m})  )  
+ (\bar\ph_m-\bar\ph^{(0)}_{m}) ^T \frac{\sum_{n=1} ^{N} \exp(\et^T_{n}\bar\ph^{(0)}_{m}) \et_{n}  } {\sum_{n=1}^{N} \exp (\et^T_{n} \bar\ph^{(0)}_{m} )},
\end{align}
where $\bar\ph^{(0)}_{m}$ is the value of $\bar\ph_m$ at the last iteration of $\ph-\gamm$ update discussed in~\algref{alg:vace}.

Taking the derivative w.r.t. $\ph_m$, we have that
\begin{align} \label{eq:gradient_phi_classify}
    \frac{\partial L_f }{\partial \bar\ph_m} \approx \sum_{n=1}^{N} y_{mn} \et_n 
    -  \frac{\sum_{n=1} ^{N} \exp(\et^T_{n} \bar\ph_m)\et_n} {\sum_{n=1}^{N} \exp (\et^T_{n} \bar\ph_m)}.
\end{align}

Note that by definition $\bar \ph_m = \nicefrac{1}{J}\sum\nolimits_{j=1}^{J} \ph_{mj}$ in the main paper, we have
\begin{align}
    \frac{\partial L_f }{\partial \ph_{mj}} = \frac{\partial L_f }{\partial \bar \ph_m} \cdot \frac{\partial \bar\ph_m }{\partial \ph_{mj}} = \frac{1}{N} \frac{\partial L_f }{\partial \bar \ph_m} = \frac{1}{N}(\sum_{n=1}^{N} y_{mn} \et_n 
    -  \frac{\sum_{n=1} ^{N} \exp(\et^T_{n} \bar\ph_m)\et_n} {\sum_{n=1}^{N} \exp (\et^T_{n} \bar\ph_m)}).
\end{align}

\textbf{Derivative of $L_s$.}
Taking the derivative of~\eqnref{eq:stable_simplified}, i.e. 
\begin{align}
    L_s   =~& \bet^T (\bar\ph_{m}\circ\bar\ph'_{m})
     - \log (\sum\nolimits_{f\in \FM} \exp(\bet^T (\bar\ph_{m}\circ\bar\ph_{f})) ),  
\end{align}
we have that 
\begin{align}
    \frac{\partial L_s }{\partial \bar\ph_m} \approx \bet^T \bar \ph'_m  - \frac{\sum_{f\in \FM}\exp (\bet^T  (\bar \ph_m \circ\bar\ph_f)  )  \bet^T \bar\ph_f} {\sum_{f\in \FM}\exp (\bet^T (\bar\ph_m \circ\bar\ph_f ))}.
\end{align}


In summary, the partial derivative of ELBO in~\eqnref{eq:full_elbo} w.r.t. $\ph_{mj}$ is 
\begin{align}
    \frac{\partial L}{\partial \ph_{mj}}
    =& \frac{\partial L_{e} }{\partial \ph_{mj}}
    +\frac{\partial L_f }{\partial \ph_{mj}}
    + \frac{\partial L_s }{\partial \ph_{mj}}.
\end{align}


Setting the derivative to $0$, we have a closed-form update rules for $\ph$ as follows:
\begin{align}\label{eq:update_phi_app}
    \ph_{mj} \propto ~& \frac{1}{\vert \Si_k\vert^{1/2}} 
      \exp[\Psi(\gamma_{mk})-\Psi(\sum\nolimits_{k'=1}^K \gamma_{k'}) 
     - \frac{1}{2}a_{mj}(\e_{mj}-\muu_k)^T\Si_k^{-1}(\e_{mj}-\muu_k)\nonumber\\ 
     + & \frac{1}{J} (\sum\nolimits_{n=1}^{N} \hat y_{ml} \et_l 
    -  \frac{\sum\nolimits_{n'=1} ^{N} \exp(\et^T_{n'} \bar\ph_{m}){\et_n}} {\sum\nolimits_{n=1}^{N} \exp (\et^T_{n} \bar\ph_m)}
     + \bet^T \bar \ph'_m  - \frac{\sum\nolimits_{f\in \FM}\exp (\bet^T  (\bar \ph_m \circ \bar\ph_f)  ) (\bet^T \bar\ph_f)} {\sum\nolimits_{f\in \FM}\exp (\bet^T (\bar\ph_m \circ \bar\ph_f ))}) ].
\end{align}
Taking derivative of~\eqnref{eq:elbo_expand} and set to $\0$, we have 
\begin{align}
    \gamma_{mk} = \alpha_k + \sum\nolimits_{j=1}^{J}  \phi_{mjk}a_{mj}.\label{eq:update_gamma_app}
\end{align}

\subsubsection{Learning} \label{app:learning}
 Similar to~\secref{app:inference}, we expand the ELBO in~\eqnref{eq:elbo}, take its derivative w.r.t. $\muu_k$ and set it to $0$:  
\begin{align}
    \frac{\partial L}{\partial \muu_k} =\sum_{m,j}\phi_{mjk}a_{mj} \Si_{k}^{-1} (\e_{mj}-\muu_k) = 0,
\end{align}
yielding the update rule for learning $\muu_k$:
\begin{align}
    \muu_k &=  \frac{\sum_{m,j}{\phi_{mjk}a_{mj} \e_{mj}}}{\sum_{m,j} \phi_{mjk}a_{mj}},\label{eq:update_mu_app}
\end{align}
where $\Si_{k}^{-1}$ is canceled out. 

Similarly, setting the derivatives w.r.t. $\Si_k$ to $\0$, i.e., 
 \begin{align}
    \frac{\partial L}{\partial \Si_k} 
    = \frac{1}{2}\sum_{m,j} \phi_{mjk}a_{mj}(-\Si_k^{-1}
    +\Si_k^{-1}(\e_{mj}-\muu_k)(\e_{mj}-\muu_k)^T\Si_k^{-1}),
 \end{align}
we have
\begin{align}
     \Si_k &= \frac{\sum_{m,j}\phi_{mjk}a_{mj} (\e_{mj}-\muu_k)(\e_{mj}-\muu_k)^T}{\sum_{m,j} \phi_{mjk}a_{mj} }. \label{eq:update_sigma_app}
\end{align}

\end{document}